\documentclass{article}

\usepackage[dvipsnames]{xcolor}
\usepackage{float}
\usepackage{microtype}
\usepackage{listings}
\usepackage{graphicx}
\usepackage{multirow}
\usepackage{booktabs} 
\usepackage{tikz}
\usepackage{amsmath}
\usepackage{amsfonts}
\usepackage{xspace}
\usepackage{etoolbox}
\usepackage[shortlabels, inline]{enumitem}
\usepackage{subcaption}
\setlist[itemize]{leftmargin=*}
\setlist[enumerate]{leftmargin=*}

\usepackage[ruled, vlined, linesnumbered]{algorithm2e}

\usepackage{hyperref}

\usepackage{tikz}
\usetikzlibrary{matrix, arrows, positioning, calc, fit, backgrounds, decorations.pathreplacing, calligraphy}

\def\commentColor{green!40!black}

\lstdefinelanguage{MLIR}{
  keywords={func, return, pmap},
  numbers=left,
  numbersep=3pt,
  xleftmargin=2em,
  escapeinside={|}{|},
  numberstyle=\tiny\color{gray},
  basicstyle=\scriptsize\ttfamily,
  columns=flexible,
  morecomment=[s][\color{\commentColor}]{/*}{*/},
  morecomment=[l][\color{\commentColor}]{//},
  mathescape=true,
}
\newcommand{\refSec}[1]{\S\ref{#1}}
\newcommand{\refFig}[1]{Figure~\ref{#1}}

\newcommand{\refAlg}[1]{Algorithm~\ref{#1}}

\newcommand{\refLine}[1]{line~\ref{#1}}

\newcommand{\device}{d}
\newcommand{\devices}{D}
\newcommand{\deviceDom}{\mathbb{D}}

\newcommand{\func}{f}
\newcommand{\funcName}{\mathit{foo}}
\newcommand{\operation}{\mathit{op}}
\newcommand{\operations}{\vec{\operation}}
\newcommand{\variable}{x}
\newcommand{\variables}{\vec{\variable}}
\newcommand{\variableB}{y}
\newcommand{\variablesB}{\vec{\variableB}}
\newcommand{\variableDom}{\mathbb{X}}

\newcommand{\opType}{O}
\newcommand{\opTypeDom}{\mathbb{O}}

\newcommand{\keywordFont}[1]{\text{\texttt{\textbf{#1}}}}

\newcommand{\callKeyword}{\keywordFont{call}}
\newcommand{\returnKeyword}{\keywordFont{return}}

\newcommand{\val}{v}
\newcommand{\vals}{\vec{\val}}
\newcommand{\valB}{w}
\newcommand{\valsB}{\vec{\valB}}

\newcommand{\absDom}{A}

\newcommand{\absSemantics}{\mathbb{S}}

\newcommand{\absState}{\rho}

\definecolor{myorange}{RGB}{254, 97, 0}
\definecolor{myblue}{RGB}{100, 143, 255}
\foreach \dColor [count=\i] in {teal, myblue!90!black, myorange, {red!90!black}, orange} {%
  \edef\d{\the\numexpr \i-1} 
  \expandafter\xdef\csname deviceColor\d\endcsname {\dColor}
} 
\newcommand{\getDeviceColor}[1]{\csname deviceColor#1\endcsname}

\newcommand{\resetTimestamps}{%
  \foreach \d in {0, 1, ..., 9} {%
    \expandafter\xdef\csname timestampOf\d\endcsname {0}}
}
\newcommand{\setTimestamp}[2]{\expandafter\xdef\csname timestampOf#1\expandafter\endcsname {#2}}
\newcommand{\getTimestamp}[1]{\csname timestampOf#1\endcsname}

\pgfkeys{
  /traceEvent/.is family, /traceEvent,
  default/.style = {device = 0, start = 0, duration = 2, color = \defaultColor}, 
  device/.estore in = \eventDevice,
  start/.estore in = \eventStart,
  duration/.estore in = \eventDuration,
  color/.estore in = \eventColor,
}

\def\eventXPadding{0.05}
\def\eventYPadding{0.1}

\newcommand{\drawEvent}[5]{
  \filldraw[draw=#4!40, fill=#4!20, rounded corners] (#2+\eventXPadding, -#1-1+\eventYPadding)
  rectangle (#3-\eventXPadding, -#1-\eventYPadding)
  node[font=\scriptsize, text centered, midway, inner sep=0pt] {#5};%
}

\newcommand{\traceEvent}[2][]{%
  \pgfkeys{/traceEvent, default, #1}%
  \pgfmathsetmacro{\startTS}{\getTimestamp{\eventDevice}}
  \pgfmathsetmacro{\endTS}{\getTimestamp{\eventDevice}+\eventDuration}
  \drawEvent{\eventDevice}{\startTS}{\endTS}{\eventColor}{#2}
  \setTimestamp{\eventDevice}{\endTS}
}

\newcommand{\commEvent}[4][]{
  \pgfkeys{/traceEvent, default, #1}%
  \pgfmathsetmacro{\maxTS}{max(\getTimestamp{#2}, \getTimestamp{#3})}
  \pgfmathsetmacro{\endTS}{\maxTS+\eventDuration}
  \filldraw[draw=\eventColor!40, fill=\eventColor!20, rounded corners]
  (\maxTS+\eventXPadding, -#3-1+\eventYPadding)
  rectangle (\endTS-\eventXPadding, -#2-\eventYPadding)
  node[font=\scriptsize, text centered, midway, inner sep=0pt] {#4};%
  \setTimestamp{#2}{\endTS}
  \setTimestamp{#3}{\endTS}
}

\pgfkeys{
  /trace/.is family, /trace,
  default/.style = {devices = 5, times = 15, scale = 1, startDevice = 0}, 
  devices/.estore in = \numDevices,
  times/.estore in = \numTimes,
  scale/.estore in = \tikzScale,
  startDevice/.estore in = \startDevice,
}

\newcommand{\drawTrace}[2][]{%
  \pgfkeys{/trace, default, #1}%

  \resetTimestamps{}
  \def\extraBit{.4}

  \begin{tikzpicture}[scale=\tikzScale, every node/.style={scale=\tikzScale}]

    \foreach \d [parse=true] in {\startDevice, ..., \numDevices-1} {
    \edef\dC{\getDeviceColor{\d}!10}
    \draw[fill=\dC, draw=\dC] (-\extraBit, -\d-1) rectangle (\numTimes, -\d);
    }

    \draw[|->, -latex] (-\extraBit, -\startDevice) -- (\numTimes, -\startDevice);
    \draw[|->, -latex] (0, -\startDevice+\extraBit) -- (0, -\numDevices-\extraBit);
    \foreach \y [parse=true] in {\startDevice, ..., \numDevices-1} {
      \node[text=\getDeviceColor{\y}] at (-.2, -\y-.5) {\y};
    }
    \node[style={font=\footnotesize}] at ({\numTimes/2}, -\startDevice+\extraBit) {Time};
    \node[style={font=\footnotesize}, anchor=south, rotate=90] at ({1*-\extraBit}, {(-\startDevice-\numDevices)/2}) {Devices};
    
    #2
  \end{tikzpicture}
}

\newcommand{\code}[1]{\textnormal{\texttt{#1}}}

\newcommand{\ir}{DistIR\xspace}

\usepackage[normalem]{ulem}

\usepackage[accepted]{mlsys2022}

\mlsystitlerunning{\ir: An Intermediate Representation and Simulator for Efficient Neural Network Distribution}

\begin{document}

\cfoot{\thepage}

\twocolumn[
\mlsystitle{\ir: An Intermediate Representation and Simulator for Efficient Neural Network Distribution}




\begin{mlsysauthorlist}
\mlsysauthor{Keshav Santhanam}{stanford}
\mlsysauthor{Siddharth Krishna}{microsoft}
\mlsysauthor{Ryota Tomioka}{microsoft}
\mlsysauthor{Tim Harris}{microsoft}
\mlsysauthor{Matei Zaharia}{stanford}
\end{mlsysauthorlist}

\mlsysaffiliation{stanford}{Stanford University}
\mlsysaffiliation{microsoft}{Microsoft}

\mlsyscorrespondingauthor{Keshav Santhanam}{keshav2@cs.stanford.edu}

\mlsyskeywords{Machine Learning, MLSys}

\vskip 0.3in

\begin{abstract}
The rapidly growing size of deep neural network (DNN) models and datasets has given rise to a variety of distribution strategies such as data, tensor-model, pipeline parallelism, and hybrid combinations thereof. Each of these strategies offers its own trade-offs and exhibits optimal performance across different models and hardware topologies. Selecting the best set of strategies for a given setup is challenging because the search space grows combinatorially, and debugging and testing on clusters is expensive.
In this work we propose \ir, an expressive intermediate representation for distributed DNN computation that is tailored for efficient analyses, such as simulation. This enables \emph{automatically} identifying the top-performing strategies without having to execute on physical hardware. Unlike prior work, \ir can naturally express many distribution strategies including pipeline parallelism with arbitrary schedules. Our evaluation on MLP training and GPT-2 inference models demonstrates how \ir and its simulator enable fast grid searches over complex distribution spaces spanning up to 1000+ configurations, reducing optimization time by an order of magnitude for certain regimes.
\end{abstract}
]



\printAffiliationsAndNotice{} 

\section{Introduction} \label{section:introduction}

Deep neural network (DNN) computation has become exponentially more expensive in recent years due to rapidly growing model and dataset sizes~\cite{rajbhandari2019zero, brown2020language, narayanan2021efficient, fedus2021switch, lepikhin2021gshard, raffel2020exploring}. As a result, distributed execution is now essential for achieving state-of-the-art machine learning performance.

This has led to a corresponding growth in the distribution strategies available for DNNs, each making different trade-offs to tailor for particular model architectures or hardware types.
For instance, data parallelism partitions input data across devices or ranks, which enables training with large batch sizes but can incur high communication costs to synchronize the copies of the model's parameters~\cite{dean2012large}.
Other strategies, such as tensor-model parallelism~\cite{shoeybi2019megatron} and pipeline parallelism~\cite{chen2012pipelined, gaunt2017ampnet, narayanan2019pipedream, huang2019gpipe} facilitate larger models but have their own drawbacks.
For example, tensor-model parallelism reduces per-GPU memory usage but requires frequent all-reduce synchronization operations which can be expensive without sufficiently fast network links.
These strategies can also be combined into hybrid strategies~\cite{krizhevsky2014one, jia2019beyond, lepikhin2021gshard, narayanan2020memory, narayanan2021efficient}, resulting in a large space of potential distribution configurations.

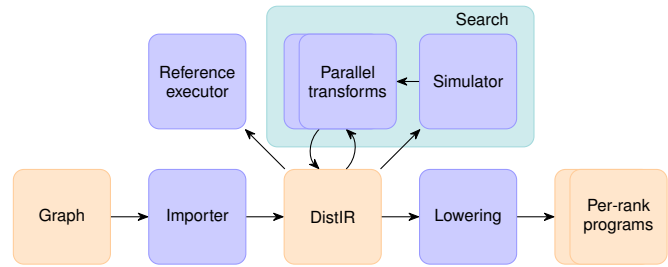
\begin{figure}[t]
  \centering
  \usetikzlibrary{arrows.meta, petri}
  \def\sysDiagScale{0.9}
  \def\ySep{4}
  \begin{tikzpicture}[%
    node distance=1.8, on grid, >={Stealth[round]}, bend angle=45, auto,
    blob/.style= {rectangle, rounded corners, minimum width=1.3cm, minimum height=1.4cm, text width=1.2cm, align=center, font=\sffamily\scriptsize},
    module/.style= {blob, draw=blue!40, fill=blue!20},
    data/.style= {blob, draw=orange!40, fill=orange!20},
    converter/.style={auto, font=\sffamily\scriptsize},
    scale=\sysDiagScale, every node/.style={scale=\sysDiagScale},
    ]
    \node[data] (graph) {Graph};
    \node[module] (importer) [right=of graph] {Importer}
    edge [<-] (graph);
    \node[data] (DistIR) [right=of importer] {\ir}
    edge [<-] (importer);
    \def\xSep{.2}
    \node[module] (transforms) [above=of DistIR] {}
    ($(transforms.south) + (-\xSep, 0)$) edge [->, bend right] ($(DistIR.north) + (-\xSep, 0)$)
    ($(transforms.south) + (\xSep, 0)$) edge [<-, bend left] ($(DistIR.north) + (\xSep, 0)$);
    \node[module] (transformsB) [right=.2cm of transforms] {Parallel transforms};
    \node[module] (simulator) [right=of transforms] {Simulator}
    edge [->] (transformsB)
    edge [<-] (DistIR);
    \node[module] (seq) [above=of importer] {Reference executor}
    edge [<-] (DistIR);
    \node[module] (lowering) [right=of DistIR] {Lowering}
    edge [<-] (DistIR);
    \node[data] (rank) [right=of lowering] {}
    edge [<-] (lowering);
    \node[data] (rankB) [right=.2cm of rank] {Per-rank programs};

    \node[anchor=east, font=\sffamily\scriptsize] at ($(simulator.north east) + (0, 6pt)$) (search) {Search};
    \def\searchColor{TealBlue}
    \begin{pgfonlayer}{background}
      \node[rounded corners, draw=\searchColor!40, fill=\searchColor!20,
      fit=(simulator)(transforms)(search), inner xsep=12, inner ysep=6, yshift=-4] (searchBox) {};
    \end{pgfonlayer}
  \end{tikzpicture}
  \caption{The workflow of optimizing distributed computation using \ir.
    We import a computation graph (e.g., ONNX) representing the DNN model, use a search algorithm with a simulator to quickly find an efficient distribution strategy, and lower the resulting computation to per-rank programs for execution.
    \ir also provides a reference executor for testing and debugging.
  }
  \label{fig-workflow-diff}
\end{figure}

How do we select the best distribution strategy for a given model and hardware configuration?
This is a challenging problem not only because of the range of strategies to choose from, but also because testing and debugging on clusters of hardware accelerators is expensive and time-consuming.

One solution is to statically analyze (e.g., simulate) the model and potential distribution strategy before execution, enabling automatic search among a set of candidate strategies for the optimal configuration.
While this approach has been shown to be promising~\cite{jia2019beyond, zhu2020daydream}, existing simulators are limited to domain-specific languages and support a limited set of distribution strategies.

To simulate a distributed computation without executing it, one needs a static representation of the program as input.
One way to build a generic simulator is to use the intermediate representation (IR) used by DNN compilers/frameworks.
However, some existing IRs are not explicit enough to simulate efficiently. 
For instance, the ONNX IR~\cite{onnx} is an operator graph that does not specify the order of execution, which means that each run of the simulator must compute a schedule.
On the other hand, DNN frameworks such as TensorFlow~\cite{abadi2016tensorflow} use a single-program-multiple-data (SPMD) style IR that can capture the schedule, but which makes it hard to express certain strategies, such as pipeline parallelism~\cite{xu2021gspmd}, as we discuss in \refSec{section:dist_ir}.


In this paper, we propose \ir, an expressive IR for distributed computation that enables efficient analysis and simulation.
By explicitly representing the global distributed computation in the IR, \ir enables quick and accurate simulation of a large range of distribution strategies including pipeline parallel hybrids.
\ir integrates with popular DNN frameworks ONNX and PyTorch, and can be extended to support other frameworks.

\ir programs have an explicit schedule, as they are ordered lists of operations as opposed to a computation graph.
\ir's semantics dictate that each device executes one operation at a time, and operations involving multiple devices execute synchronously on all participating devices (e.g., \code{Send} blocks both sender and receiver).\footnote{Finer-grained asynchronous concurrent operations can be modeled in terms of these primitive synchronous operations.}

\ir is expressive, and can represent a diverse range of distribution strategies. For example, it supports data and tensor-model parallelism, as well as hybrid strategies involving pipeline parallelism---which other systems do not support~\cite{jia2019beyond} or support in a restricted form~\cite{xu2021gspmd}.

We build a framework of analyses for \ir programs, including simulation.
\ir's distributed semantics allows us to combine analytic or empirical cost models for each operator in order to simulate the distributed computation, handling synchronization accurately.
We use a mixed concrete/abstract execution in order to infer the shapes of inputs to each operation, while supporting dynamic operators such as \code{Reshape}.
We also implement a reference executor that aids the development of new strategies, and a lowering of \ir to PyTorch that enables running the distributed computation on GPUs.

Our evaluation demonstrates that \ir and its simulator can analyze distributed performance at scale and can automatically optimize models to quickly find efficient distribution strategies. We show this by conducting simulated grid searches over complex spaces (up to 1000+ configurations) generated by applying a D/T/P transform~\cite{rasley2020deepspeed, narayanan2021efficient}, which combines data, tensor-model, and pipeline parallelism, to MLP training and GPT-2 inference computations. Our simulator reduces search time by an order of magnitude. We also verify that the simulator accurately ranks distribution strategies with respect to their true performance on real hardware. Finally, we show that simulation time scales linearly with the number of operations.

\ir has a few additional benefits.
For instance, distribution strategies are be implemented as IR-to-IR transformations in \ir.
This allows separating the distribution strategy from DNN model definitions and into a library of reusable distributions.
Writing a new distribution strategy is also simplified by the fact that one can reuse the lowering pass that produces the per-rank programs (the low-level code executed on each device).

\ir can also easily be extended with new primitive operators by providing definitions and cost models for them.
In this paper, we instantiate \ir with ONNX primitives and MPI communication primitives.
Frameworks like XLA or JAX~\cite{leary2017xla, jax2018github} can be supported by instantiating \ir with their primitive operators and lowering to their respective backends.

In summary, this paper makes the following contributions:
\begin{itemize}
\item We present \ir, an explicit, expressive, and extensible IR for distributed computation (\refSec{section:dist_ir}). Our implementation contains an ONNX importer and a lowering to PyTorch to support running state-of-the-art models on GPUs.
\item We build an abstract execution framework  (\refSec{section:analyses}) over \ir that enables various analyses, such as efficient simulation. Our simulator uses mixed concrete and abstract execution to accurately predict the cost of computations involving dynamic operations such as \code{Reshape}.
\item We demonstrate how \ir facilitates optimizing distributed computation by applying a grid search algorithm over the D/T/P space of distributions for training of MLP models and inference with GPT-2 models. By using \ir's simulator, we are able to identify competitive distribution strategies ahead-of-time in a few hours, compared to the days it would take to try all possible strategies.
\end{itemize}



\section{\ir} \label{section:dist_ir}

In this section we define the \ir language and semantics, and discuss its design and expressivity.

\begin{figure}[t]
\begin{lstlisting}
func @dense(%w, %x) {
  %A, %b = UnpackTuple(%w)
  %h = Gemm(%x, %A, %b)
  %a = Relu(%h)
  return %a
}

func @denseGrad(%wb, %x, %da) { ... }

func @lossGrad(%p, %y) { ... }
\end{lstlisting}\vspace{-8pt}
\begin{lstlisting}[%
  emph={[1]D1, x, w1, x_1, as_1, x_2, as_2, dar_1, dw1_1, dar_2, dw1_2, dw1, w1_new},
  emphstyle={[1]\color{\getDeviceColor{1}}},
  emph={[2]D2, y, w2, y_1, ar_1, y_2, ar_2, das_1, dw2_1, das_2, dw2_2, dw2, w2_new, p_1, p_2, dp_1, dp_2},
  emphstyle={[2]\color{\getDeviceColor{2}}},
  firstnumber=last,
  % moredelim=**[is][\color{\getDeviceColor{1}}]{|}{|},
  % moredelim=**[is][\color{\getDeviceColor{2}}]{|(}{)|},
  ]

func @mlpPP(%w1: D1, %w2: D2, %x: D1, %y: D2) {
  // Split into microbatches 1 and 2
  %x_1: D1, %x_2: D1 = Split(%x, dim=0, num_splits=2)
  %y_1: D2, %y_2: D2 = Split(%y, dim=0, num_splits=2)
  // Pipeline
  %as_1: D1 = @dense(%w1, %x_1)
  %ar_1: D2 = Send(%as_1, 2)
  %as_2: D1 = @dense(%w1, %x_2)  |\label{line-F1_2}|
  %p_1: D2 = @dense(%w2, %ar_1)  |\label{line-F2_1}|
  %ar_2: D2 = Send(%as_2, 2)        |\label{line-S1_2}|
  %dp_1: D2 = @lossGrad(%p_1, %y_1)
  %dw2_1, %das_1: D2 = @denseGrad(%w2, %ar_1, %dp_1)
  %dar_1: D1 = Send(%das_1, 1)
  %dw1_1: D1, _ = @denseGrad(%w1, %x_1, %dar_1)
  %p_2: D2 = @dense(%w2, %ar_2)
  %dp_2: D2 = @lossGrad(%p_2, %y_2)
  %dw2_2, %das_2: D2 = @denseGrad(%w2, %ar_2, %dp_2)
  %dar_2: D1 = Send(%das_2, 1)
  %dw1_2: D1, _ = @denseGrad(%w1, %x_2, %dar_2)
  // Weight update (WU)
  %dw1: D1 = Sum(%dw1_1, %dw1_2)
  %w1_new: D1 = Optimizer(%w1, %dw1)
  %dw2: D2 = Sum(%dw2_1, %dw2_2)
  %w2_new: D2 = Optimizer(%w2, %dw2)
  return %w1_new, %w2_new
}

\end{lstlisting}
\caption{\ir code listing for pipeline-parallel training of a 2-layer MLP model over 2 devices.
  We use compiler naming conventions in code listings, e.g. \code{\%x} for variables and \code{@foo} for functions.
  The functions \code{@denseGrad} implements the backwards pass for an MLP layer, and \code{@lossGrad(\%p, \%y)} computes the gradient of the predictions \code{\%p} given the labels \code{\%y}.
  In \ir, \code{Send} encapsulates both sending and receiving.
  We annotate variables with \textcolor{\getDeviceColor{1}}{\code{D1}} and \textcolor{\getDeviceColor{2}}{\code{D2}}, and color them blue and orange, to represent that they live on device 1 and 2 respectively.
}
  \label{fig-pipeline-parallel}
\end{figure}

\begin{figure}[t]
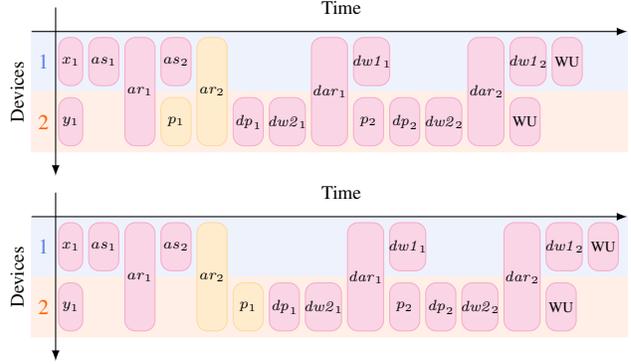

  \def\defaultColor{magenta}
  \def\highlightColor{yellow!30!orange}
  \def\shortDur{.5}
  \def\medDur{.6}
  \def\longDur{.7}

  \centering
  \newcommand{\grad}[1]{\mathit{d#1}}
  \drawTrace[startDevice=1, devices=3, times=9.5, scale=0.8]{
    \traceEvent[device=1, duration=\shortDur]{$x_1$}
    \traceEvent[device=2, duration=\shortDur]{$y_1$}
    \traceEvent[device=1, duration=\medDur]{$as_1$}
    \commEvent[duration=\medDur]{1}{2}{$ar_1$}
    \traceEvent[device=1, duration=\medDur]{$as_2$}
    \traceEvent[device=2, duration=\medDur, color=\highlightColor]{$p_1$}
    \commEvent[duration=\medDur, color=\highlightColor]{1}{2}{$ar_2$}
    \traceEvent[device=2, duration=\medDur]{$\grad{p}_1$}
    \traceEvent[device=2, duration=\longDur]{$\grad{w2}_1$}
    \commEvent[duration=\longDur]{1}{2}{$\grad{ar}_1$}
    \traceEvent[device=1, duration=\longDur]{$\grad{w1}_1$}
    \traceEvent[device=2, duration=\medDur]{$p_2$}
    \traceEvent[device=2, duration=\medDur]{$\grad{p}_2$}
    \traceEvent[device=2, duration=\longDur]{$\grad{w2}_2$}
    \commEvent[duration=\longDur]{1}{2}{$\grad{ar}_2$}
    \traceEvent[device=1, duration=\longDur]{$\grad{w1}_2$}
    \traceEvent[device=1, duration=\medDur]{WU}
    \traceEvent[device=2, duration=\medDur]{WU}
  }
  \\
  \drawTrace[startDevice=1, devices=3, times=9.5, scale=0.8]{
    \traceEvent[device=1, duration=\shortDur]{$x_1$}
    \traceEvent[device=2, duration=\shortDur]{$y_1$}
    \traceEvent[device=1, duration=\medDur]{$as_1$}
    \commEvent[duration=\medDur]{1}{2}{$ar_1$}
    \traceEvent[device=1, duration=\medDur]{$as_2$}
    \commEvent[duration=\medDur, color=\highlightColor]{1}{2}{$ar_2$}
    \traceEvent[device=2, duration=\medDur, color=\highlightColor]{$p_1$}
    \traceEvent[device=2, duration=\medDur]{$\grad{p}_1$}
    \traceEvent[device=2, duration=\longDur]{$\grad{w2}_1$}
    \commEvent[duration=\longDur]{1}{2}{$\grad{ar}_1$}
    \traceEvent[device=1, duration=\longDur]{$\grad{w1}_1$}
    \traceEvent[device=2, duration=\medDur]{$p_2$}
    \traceEvent[device=2, duration=\medDur]{$\grad{p}_2$}
    \traceEvent[device=2, duration=\longDur]{$\grad{w2}_2$}
    \commEvent[duration=\longDur]{1}{2}{$\grad{ar}_2$}
    \traceEvent[device=1, duration=\longDur]{$\grad{w1}_2$}
    \traceEvent[device=1, duration=\medDur]{WU}
    \traceEvent[device=2, duration=\medDur]{WU}
  }

  \caption{Traces (not-to-scale) for \code{@mlpPP} from \refFig{fig-pipeline-parallel} (top) and the (slower) program obtained by swapping lines \ref{line-F2_1} ($p_1$) and \ref{line-S1_2} ($ar_2$) of \code{@mlpPP}.
    Each op is labelled by its (first) return value (with \%s omitted) and ``WU'' represents the weight update.
  }
  \label{fig-pipeline-trace}
\end{figure}


\ir is an intermediate representation (IR) for distributed computation based on the static single assignment (SSA) form.
The top-level container is a module, which is comprised of a sequence of functions.
A function consists of a name, a sequence of variables that are function parameters, and a sequence of operations that make up the function body.
Operations come in three forms: invocations to a primitive operation (henceforth op), calls to other functions defined in the same module, or return statements.
\refFig{fig-pipeline-parallel} shows an example \ir program.

\ir is designed to be extensible by being parametric on the set of primitive op types $\opTypeDom$.
The core framework requires only that ops be registered along with their function signatures.
(The simulator in \refSec{section:simulator} requires abstract implementations and cost functions for each registered op.)
\ir's type system also allows extension with new types as required (we omit type annotations in our listings for brevity).
We have instantiated \ir with ONNX ops, corresponding gradient ops, and MPI communication ops.

All programs in \ir are essentially straight-line code: there are no loops, branches, or recursive function calls.
However, note that primitive ops can abstract arbitrarily complex computations, including on multiple devices, as long as we can define cost models for them.

For example, consider the program to train a 2-layer multi-layer perceptron (MLP) model over 2 devices using a pipeline parallel strategy (\refFig{fig-pipeline-parallel}).
The function \code{@dense} represents a single layer in an MLP model, and uses primitive ops \code{Gemm} and \code{Relu} from the ONNX standard, and an \code{UnpackTuple} primitive to unpack a tuple of weights (for brevity).
The \code{@mlpPP} function splits the training data into two microbatches and then executes the forward pass and backward pass on each microbatch before summing up the gradients and updating the weights.
The code for each microbatch is interleaved in order to capture the efficient pipelined execution shown in the trace in \refFig{fig-pipeline-trace}, as explained in the next section.

\subsection{Distributed Semantics}



\ir programs execute on a distributed computation model over a finite fixed set of devices $\deviceDom$, each of which is single-threaded and can execute at most one operation at a time.
Each operation executes in a synchronous manner on a set of devices $\devices \subseteq \deviceDom$.
This means that execution of the op waits until all the involved devices are free before proceeding.
This set of devices can depend on the runtime input values and their locations, e.g. a \code{Send(\%x, 2)} will run on devices 1 and 2 if its input \code{\%x} lives on device 1.\footnote{Since \ir models the global computation over all devices, there is no need to have separate send and receive ops.}
The op register contains this information, along with the concrete implementations of each primitive op in $\opTypeDom$.

\ir has an explicit \emph{schedule}: operations execute in the program order, but consecutive operations on disjoint sets of devices execute in parallel.
For example, consider \code{@mlpPP} from \refFig{fig-pipeline-parallel}.
Assuming its input values \code{\%w1} and \code{\%x} (respectively, \code{\%w2} and \code{\%y}) live on device 1 (respectively, 2), then the first two \code{Split} ops execute in parallel on devices 1 and 2 (\refFig{fig-pipeline-trace}, top).
After this, the \code{@dense} returning \code{\%as\_1} and the \code{Send} returning \code{\%ar\_1} execute in sequence (because they both involve device 1), followed by simultaneous computation of \code{\%as\_2} and \code{\%p\_1} (because they involve separate devices).

However, if we swapped lines \ref{line-F2_1} (\code{\%p\_1}) and \ref{line-S1_2} (\code{\%ar\_2}), then because the \code{Send} involves both devices, it blocks \code{\%p\_1} on device 2 from executing until it completes (\refFig{fig-pipeline-trace}, bottom).
We see that \ir enforces the schedule given by program order, regardless of the fact that \refLine{line-F2_1} and \refLine{line-S1_2} have no data dependencies and can be swapped without changing the program's return value.

\ir's representation of pipeline training (\code{@mlpPP}) captures the distributed computation on all devices in the same function.
It captures the way the inputs are split into microbatches in the first few lines; the way the model is partitioned into multiple stages using the \code{@dense} function; and the pipeline schedule that determines the order in which microbatches execute on a device in the program order of the multiple calls to \code{@dense}.


Note that we do not expect users to write \ir code manually.
Users can continue writing forward-only code (e.g., \code{@dense}) in a frontend like PyTorch (which can also generate the backwards pass, e.g \code{@denseGrad}) and export it to ONNX or XLA, from which we import to \ir.
\ir then distributes the code by applying transforms (resulting in, e.g, \code{@mlpPP}).
The verbose nature of \ir makes it easy to perform distribution, and to analyze and simulate the resulting programs.

\paragraph{Comparison to SPMD.}
Single-program-multiple-data (SPMD) representations struggle with computations such as \code{@mlpPP} because, as can be seen from \refFig{fig-pipeline-trace}, each device executes a different program.
GSPMD~\cite{xu2021gspmd} uses wrapper code (some form of a vectorized map) \emph{outside} of the IR to achieve pipelining.
This makes it hard to simulate as the IR does not specify key details like the pipeline schedule.
Moreover, the SPMD restriction limits this encoding to partitions where each device executes the same computation, which rules out, e.g., language models that start and end with embedding layers.
One could explore encoding pipeline parallelism by extending the IR with branching ops (imagine a program that branched on the rank and executed either the blue or orange lines of \code{@mlpPP}), but our representation is arguably more natural.





\subsection{Expressivity}
\label{section:expressivity}

\ir is expressive enough to represent many distributed DNN training strategies of interest, including data parallelism, tensor-model parallelism~\cite{shoeybi2019megatron}, multiple pipeline-parallel schedules, and hybrid combinations of these strategies~\cite{krizhevsky2014one, rasley2020deepspeed}, as demonstrated in \refSec{section:evaluation}. Since \ir is designed to be a generic distributed programming language it can also express state-of-the-art techniques such as gradient checkpointing~\cite{chen2016training} and ZeRO partitioning~\cite{rajbhandari2019zero}; we provide examples of these in the Appendix (Figures~\ref{fig:gradient_checkpointing} and~\ref{fig:zero} respectively). 


\paragraph{Limitations.} \ir's explicit design means that some computations are harder to model.
For example, the assumption that each primitive op in \ir is blocking synchronous means one must use lower-level communication primitives such as \code{Send} to model the behavior of fine-grained collective communication algorithms where some devices perform useful work before others are ready.
Another common optimization is to overlap communication with computation on devices with multiple streams.
Expressing this in \ir needs a more verbose approach of using a \ir device per stream, and specifying that devices representing streams within the same GPU have low or zero communication cost (see \refSec{section:simulator}).

\section{Analyses}
\label{section:analyses}

This section presents an analysis framework, based on abstract interpretation~\cite{cousot1977abstract, cousot1979systematic}, that we use to build a reference executor, a type (and shape) propagator, a PyTorch backend, and a simulator to estimate the runtime and memory consumption of \ir programs.


At a high-level, abstract interpretation can be thought of as interpreting a \ir program line-by-line, but with a state that maps each variable to an abstract value (such as the type \code{Int}) instead of a concrete value (such as \code{42}).
These abstract values represent the set of possible values that the variable can have over all executions of the program.

Abstract interpreters are parametric on the \emph{abstract domain}, which consists of a set $\absDom$ of abstract values and an abstract semantics $\absSemantics$.
The latter defines abstract implementations of primitive ops over this domain, represented as a mapping from op type $\opType \in \opTypeDom$ to a function $\absSemantics[\opType] \colon \absDom^n \to \absDom^m$ over abstract values.
For abstract interpretation to be sound, the abstract semantics must abstract the concrete semantics (more details in~\cite{cousot1977abstract, cousot1979systematic}).

\SetAlFnt{\small}

\begin{algorithm}[t]
\caption{An Abstract Interpreter for \ir}
\label{alg-simulator}

\SetKwFor{ForEach}{foreach}{}{end foreach}%
\SetKwSwitch{Switch}{Case}{Other}{switch}{}{case}{otherwise}{end case}{endsw}%

\LinesNumberedHidden
\DontPrintSemicolon
\SetKwInOut{Given}{given}\SetKwInOut{Input}{inputs}\SetKwInOut{Output}{outputs}

\Given{an abstract domain $(\absDom, \absSemantics)$}
\BlankLine
\Input{a function $\func(\variables)$ and a list of input values $\vals$}
\Output{the final abstract state $\absState$}
\BlankLine
$\absState \gets $ new abstract state mapping $\variables$ to $\vals$\;
\ForEach{$\operation \in \operations$}{
  \uCase{$\operation$ \textnormal{is} $\variablesB = \opType(\variables)$}{
    $\valsB \gets$ run abstract semantics $\absSemantics[\opType](\absState(\variables))$\;
    $\absState \gets$ update $\variablesB$ to $\valsB$\;
  }
  \uCase{$\operation$ \textnormal{is} $\variablesB = \callKeyword \; \funcName(\variables)$}{
    $\absState' \gets $ call Abstract Interpreter on $\funcName$ and $\absState(\variables)$\;
    $\absState \gets$ update with $\variablesB$ from $\absState'$\;
  }
  \Case{$\operation$ \textnormal{is} $\returnKeyword \; \variables$}{
    \Return $\absState$;
  }
}
\end{algorithm}

\refAlg{alg-simulator} gives the algorithm for abstract interpretation of a \ir function $\func$ on a list of input (abstract) values $\vals$.
It begins by creating an abstract state $\absState \colon \variableDom \to \absDom$ that maps the formal parameters $\variables$ to the given arguments $\vals$.
It then proceeds operation by operation: for a regular op $\opType$ it looks up the semantics and runs it on the arguments as given by $\absState$; for function calls it recursively calls the abstract interpreter on the function and appropriate argument values; and for return statements it returns the final abstract state.

An example instantiation of abstract interpretation is type propagation.
The abstract domain consists of primitive types tagged with device ID (e.g., \code{Int32[0]}) and abstract tensors, which are tuples of data type, shape, and device (e.g., \code{Tensor[Float16, (128, 64), 1]}). The abstract implementation of each op checks that the op's inputs match the expected types and returns the type(s) of the output(s).

\begin{figure*}[t]
    \centering
\begin{minipage}[t]{.32\linewidth}
\begin{lstlisting}[numbers=none]
...

// %3114 is an input
%211 = Add(...)
%218 = Shape(%211)
%219 = Constant[value = 2]()
%220 = Gather[axis = 0](%218, %219)
%223 = Unsqueeze[axes = [0]](%220)
%224 = Concat[axis = 0](%3114, %223)
%225 = Reshape(%211, %224)
%226 = Gemm(%225, ...)
...
\end{lstlisting}
\end{minipage}
\begin{minipage}[t]{.32\linewidth}
\begin{lstlisting}[numbers=none]
// Abstract (only) interpretation:
// ---------------------------------
// Tensor[Int32, (1,), 0]
// Tensor[Float32, (256, 8, 768), 0]
// Tensor[Int32, (3,), 0]
// Int32
// Int32
// Tensor[Int32, (1,), 0]
// Tensor[Int32, (2,), 0]
// Tensor[Float32, ??, 0]
\end{lstlisting}
\end{minipage}
\begin{minipage}[t]{.3\linewidth}
\begin{lstlisting}[numbers=none]
// Mixed (abs/conc) interpretation:
//---------------------------------
// [-1]
// Tensor[Float32, (256, 8, 768), 0]
// [256, 8, 768]
// 2
// 768
// [768]
// [-1, 768]
// Tensor[Float32, (2048, 768), 0]
\end{lstlisting}
\end{minipage}
    \caption{A snippet from the GPT-2 model that requires mixed abstract/concrete interpretation in order to simulate accurately. The shape of the input of \code{Gemm} depends on the concrete values of \code{\%3114} and \code{\%218}.}
    \label{fig-reshape-example}
\end{figure*}

\subsection{Reference Executor}\label{section:sequential-executor}

We implement a reference sequential executor as an instantiation of our framework.
This is used to check the output of distributed \ir programs without executing them on a cluster, which helps develop and debug distribution strategies.
We use an abstract domain consisting of concrete values (technically, each value represents a singleton set) and abstract implementations of each op perform a sequential version of its computation.
For example, an \code{MPIGather} op concatenates its inputs on the specified axis.


\subsection{Lowering and PyTorch backend}\label{section:lowering}

We also perform a device placement analysis using the abstract interpreter to perform the lowering from a \ir program representing a distributed computation to the per-rank program executed by each participating device.
We reuse the type abstract domain, as each type is tagged with device information. The abstract implementation of each op checks that the input values live on the expected devices and then returns abstract values corresponding to the devices on which each output resides.
For instance, the implementation of \code{MatMul} checks that all inputs are on a single device $\device$ and returns an abstract tensor on device $\device$, whereas an \code{Allreduce} checks that inputs are on distinct devices and returns tensors on the same list of devices.

After interpretation, we project the input program to every device $\device$ by filtering out all ops without inputs or outputs on $\device$.
We take the resulting per-rank programs and execute them using PyTorch by mapping each \ir op to the corresponding PyTorch implementation.
We use Python's \code{multiprocessing} library to spawn a process for each rank, and maintain a mapping from ranks to GPUs/CPUs.

\subsection{Simulator}\label{section:simulator}

The main application of our abstract interpreter is a \ir program simulator that can estimate its runtime and memory consumption.

Our simulator works on the principle that, given the runtime and memory consumption of each op, the execution of a distributed program is determined by the order in which ops are executed.
Since \ir fixes the op schedule in the IR, the problem reduces to simulating the execution of each op.

We assume that the (runtime and memory) cost of each op can be modeled by cost functions that depend only on the shapes of its tensor inputs.
In order to find the shapes of intermediate values, we abstractly interpret the program using the type abstract domain (recall that our tensor types contain shape information), and abstract implementations that propagate shape information.
E.g., an (elementwise) \code{Add} on a pair of identical abstract tensors $(t,  t)$ will return $t$.
We then build the execution trace in a second pass that estimates the runtime of each op using its cost function on the input shapes  (and accounts for synchronizing ops accordingly).
For example, \code{Add}'s cost function on $(t, t)$ returns $N / f$, where $N$ is the number of elements in $t$ and $f$ is the device performance (flop/s).

A big challenge to accurate simulation is the use of dynamic ops such as \code{Shape}, where the output shape (and hence the cost of downstream ops) depends on the concrete value of the inputs.
For example, consider the \ir snippet in \refFig{fig-reshape-example}, taken from the GPT-2 model.
This code dynamically reshapes tensor \code{\%211} from a rank 3 tensor (e.g., of shape  $(256, 8, 768)$) to a rank 2 tensor (e.g.,  $(2048, 768)$).

However, if we perform an abstract interpretation of this code using only abstract values (as shown in the middle column), then \code{\%218} and other downstream variables will have only shape information.
In particular, we will not know the value of the second argument to \code{Reshape}, which means we cannot deduce the shape of the output \code{\%225}.
In turn, this means we cannot simulate the \code{Gemm} op at the end.

We solve this issue by using a mixed abstract domain containing both concrete values (e.g., 11, -4.56, [1, 2, 3]), as well as the abstract types defined above.
By interpreting the program on an abstract tensor value for \code{\%211} but a concrete value for \code{\%3114} (as shown in the right column of \refFig{fig-reshape-example}) we obtain the correct shape for the output of the \code{Reshape} op, and are able to simulate the \code{Gemm} op successfully.

Supporting such mixed interpretation requires the semantics of the interpreter to contain both abstract and concrete (or mixed) implementations.
For ops such as \code{Shape}, we add implementations that convert an abstract input like \code{Tensor[Float32, (128, 64), 0]} to the concrete output \code{[128, 64]}.
An op such as \code{Reshape} can work on either an abstract or concrete first argument, but requires a concrete value for the second, and returns an appropriately reshaped value.
As it is useful to support multiple implementations for each op based on whether the inputs are abstract or concrete, we implement a dynamic dispatch algorithm in the interpreter that picks the most precise matching implementation for the given op and argument values.

By carefully choosing which input values to abstract and which to remain concrete we can quickly yet accurately estimate the runtime of tensor ops.
We also estimate the live memory profile for each device by calculating the memory requirement of each tensor from its shape and assuming that it is live from the time it is created until its last usage.




\section{Implementation} \label{section:implementation}

\ir is implemented in roughly 8500 lines of Python code. The code is organized into components for the representation itself (800 LoC), analysis passes such as simulation and reference execution (2500 LoC), parallel IR-to-IR transforms (1800 LoC), the PyTorch backend (600 LoC), and example models / grid search infrastructure (2800 LoC). 

Our simulator implementation uses a combination of simple analytic cost functions (e.g., for elementwise ops) and empirical cost functions (e.g., for \code{MatMul} and \code{AllReduce}) for op runtimes. The latter are linear regression models in terms of the sizes of the input tensors.
We calibrate the simulator by fitting these regression models on microbenchmarks where we run a single op on inputs of various sizes.
Some of the regression coefficients correspond to hardware parameters such as GPU DRAM bandwidth, kernel launch overhead, and network bandwidths.

We implemented two example model architectures for demonstrating the utility of \ir: a GPT-2 example for inference, and a synthetic MLP example for training. The GPT-2 example is derived from the HuggingFace GPT-2 implementation via an ONNX sample model, and we create the synthetic MLP models directly in DistIR.

We also implemented a D/T/P parallel transform for each example model in order to enumerate and search through the space of possible distributed strategies. The transforms take as input a sequential \ir program, as well as the data-parallel degree $D$, the tensor-model-parallel degree $T$, the pipeline-parallel degree $P$, and the number of microbatches $K$. The transforms then return a new program representing the appropriately distributed computation. For pipeline parallelism we uniformly partition the model into the $P$ pipeline parallel stages and apply synchronous 1F1B scheduling~\cite{narayanan2019pipedream}, but our pipeline-parallel implementation can easily be adapted to non-uniform partitioning strategies or different schedules.
\section{Evaluation} \label{section:evaluation}


\begin{table*}[ht]
\resizebox{\linewidth}{!}{
\begin{tabular}{@{}llrrrrrrrr@{}}
\toprule
\multicolumn{1}{c}{Workload} & \multicolumn{1}{c}{Model} & \multicolumn{1}{c}{$n_{\text{layer}}$} & \multicolumn{1}{c}{$d_{\text{model}}$} & \multicolumn{1}{c}{\begin{tabular}[c]{@{}c@{}}Model Size\\ (GB)\end{tabular}} & \multicolumn{1}{c}{$N_{\text{grid}}$} & \multicolumn{1}{c}{\begin{tabular}[c]{@{}c@{}}Average\\ real trial\\ (minutes)\end{tabular}} & \multicolumn{1}{c}{\begin{tabular}[c]{@{}c@{}}Real\\  grid search\\ (minutes)\end{tabular}} & \multicolumn{1}{c}{\begin{tabular}[c]{@{}c@{}}Simulated\\ grid search\\ (minutes)\end{tabular}} & \multicolumn{1}{c}{\begin{tabular}[c]{@{}c@{}}\ir\\ optimization\\ (minutes)\end{tabular}} \\ \midrule
\multirow{3}{*}{Training}    & MLP 1B                    & 16                           & 8192                         & 2.2                                                                           & \multirow{3}{*}{75}         & 2                                                                                            & 150                                                                                                   & \textless 1                                                                                     & \textbf{20}                                                                                                      \\
                             & MLP 17B                   & 64                           & 16384                        & 34.4                                                                          &                             & 6                                                                                            & 450                                                                                                   & 1                                                                                               & \textbf{61}                                                                                                      \\
                             & MLP 103B                  & 96                           & 32768                        & 206.2                                                                         &                             & 19                                                                                           & 1425                                                                                                  & 2                                                                                               & \textbf{192}                                                                                                     \\ \midrule
\multirow{3}{*}{Inference}   & GPT-2 1.6B                & 24                           & 2048                         & 3.2                                                                           & \multirow{3}{*}{1035}       & 3                                                                                            & 3105                                                                                                  & 35                                                                                              & \textbf{65}                                                                                                      \\
                             & GPT-2 13B                 & 40                           & 5140                         & 25.8                                                                          &                             & 6                                                                                            & 6210                                                                                                  & 58                                                                                              & \textbf{118}                                                                                                    \\
                             & GPT-2 175B                & 96                           & 12288                        & 394.4                                                                         &                             & 19                                                                                           & 19665                                                                                                 & 138                                                                                             & \textbf{328}                                                                                                     \\ \bottomrule
\end{tabular}
}
\caption{The models used in our evaluation and a comparison of \ir optimization time versus exhaustive search.
Model sizes are computed assuming 16-bit floating point precision. $N_{\text{grid}}$ refers to the number of grid search configurations per model; note that for training we fix a particular batch size whereas for inference we treat the batch size as a free variable.
\ir optimization involves simulating the performance of all $N_{\text{grid}}$ configurations and then trying the top 10 predictions on actual hardware.
We estimate the time for exhaustive grid search on hardware by multiplying the average time for running a configuration on physical hardware with $N_{\text{grid}}$.
}
\label{table:large_scale_configs}
\end{table*}

Our evaluation demonstrates the following key results:
\begin{enumerate}
    \item \ir can be used to identify efficient distribution strategies up to an order of magnitude faster than exhaustive manual exploration on physical hardware, including strategies that are not covered by existing systems for distributed optimization. (\refSec{section:automated_distribution})
    \item The \ir simulator accurately reflects the relative ranking of distribution strategies. (\refSec{section:simulator_accuracy})
    \item The \ir simulator scales linearly with respect to the program op count. (\refSec{section:scalability})
\end{enumerate}


We evaluate these claims on six model architectures split across training and inference workloads, as specified in Table~\ref{table:large_scale_configs}. For brevity we refer to these models as MLP 1B, MLP 17B, etc., combining the model name and parameter count. We note that the 175 billion parameter GPT-2 model is similar to the largest model evaluated in~\cite{brown2020language}, i.e. the canonical GPT-3.~\footnote{While the model sizes we select match up exactly with the parameter counts from~\cite{brown2020language}, we use a GPT-2 architecture as opposed to GPT-3. However, the architectural differences are minor, as explained in~\cite{brown2020language}.}

We run all experiments evaluating our \ir PyTorch backend on an NVIDIA DGX-2 node with 16 V100 GPUs, each with 32 GB of memory and connected via NVLink. For experiments evaluating our simulator we use a 56-core, 2.60 Ghz Intel Xeon Gold 6132 CPU. We calibrated the simulator's parameters to the DGX machine specifications using the procedure described in \refSec{section:implementation}.

\subsection{Distributed Search Space} \label{section:search_space}

\begin{figure*}[ht]
    \centering
    \includegraphics[width=0.75\linewidth]{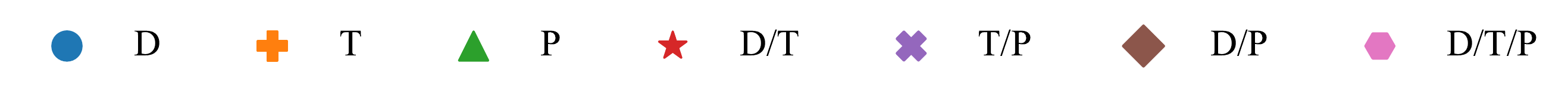} \\
    \begin{subfigure}[b]{0.30\linewidth}
        \centering
        \includegraphics[width=1.0\columnwidth]{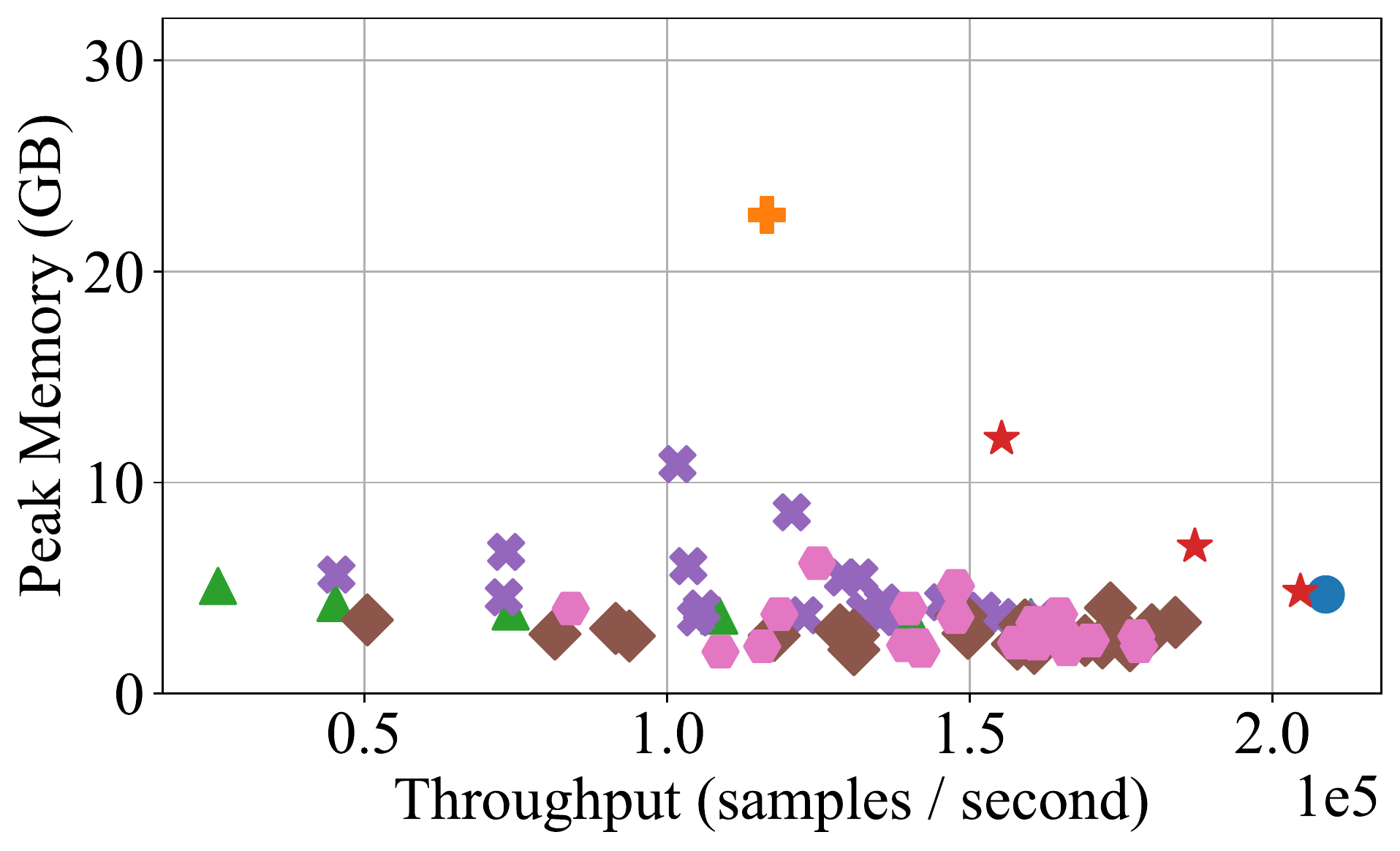}
        \caption{
        \small MLP 1B, batch size 65536}
    \end{subfigure}
    \begin{subfigure}[b]{0.30\linewidth}
        \centering
        \includegraphics[width=1.0\columnwidth]{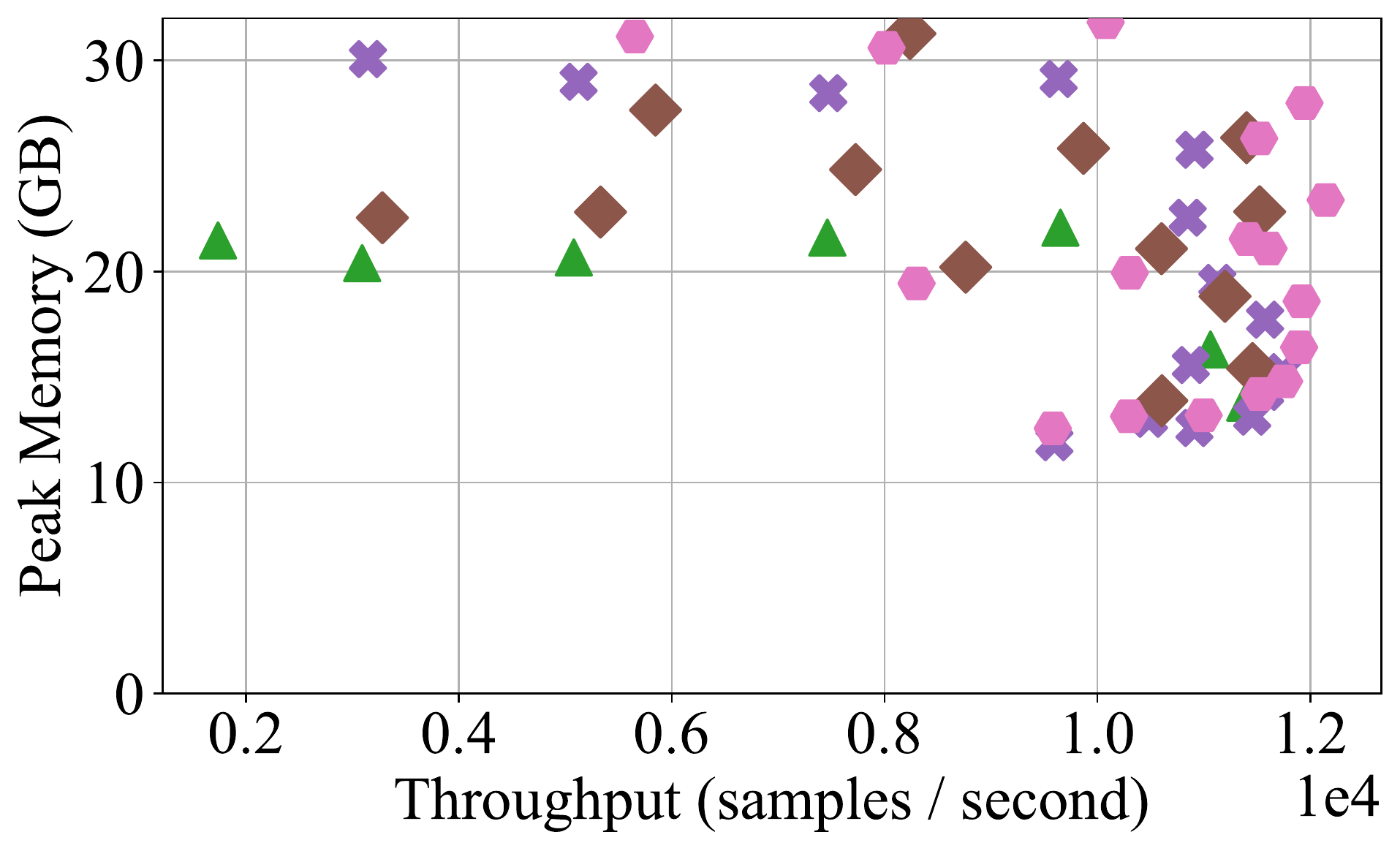}
        \caption{
        \label{fig:mlp_memory_vs_throughput_17B}
        \small MLP 17B, batch size 65536
        }
    \end{subfigure}\begin{subfigure}[b]{0.30\linewidth}
        \centering
        \includegraphics[width=1.0\columnwidth]{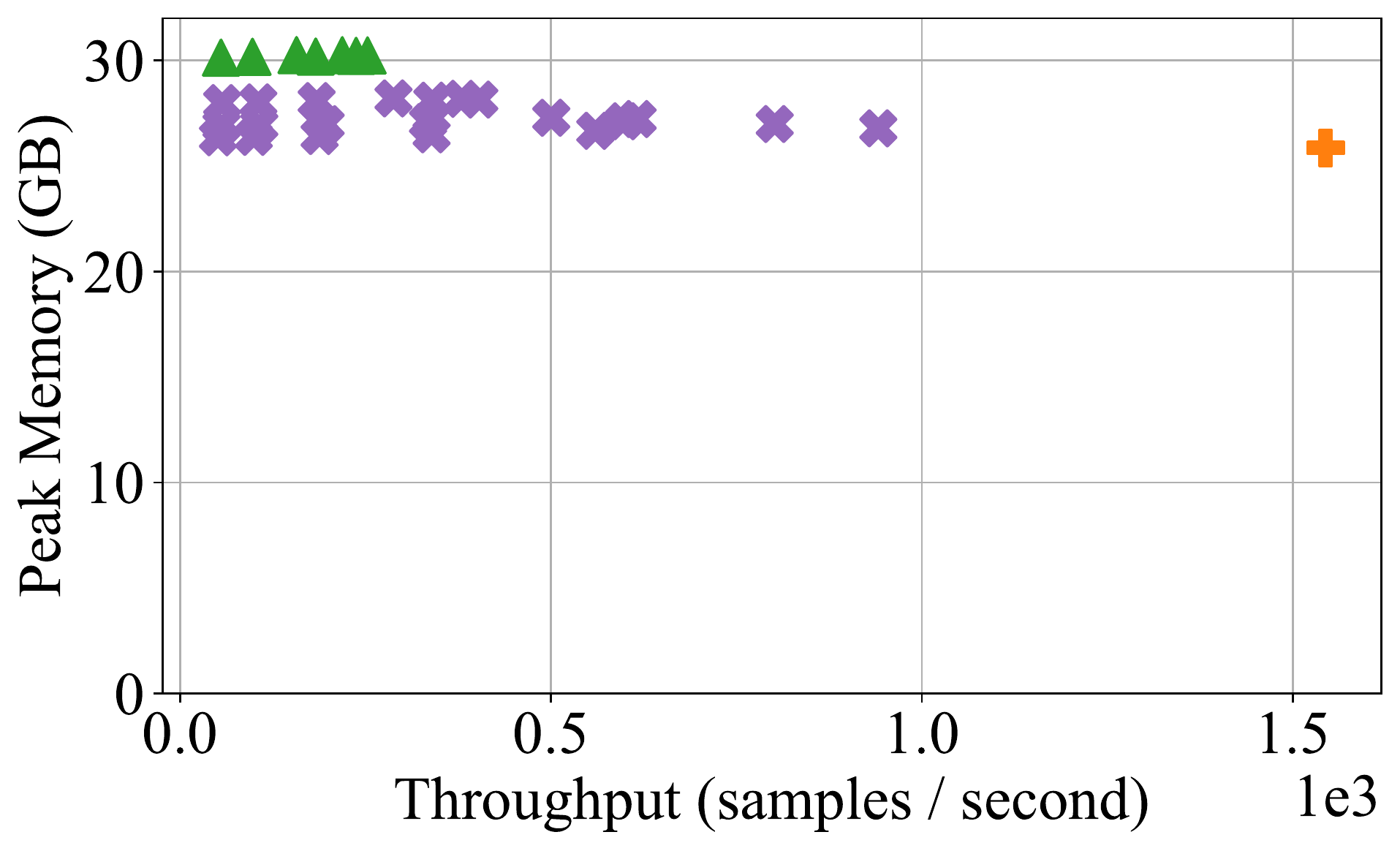}
        \caption{
        \label{fig:mlp_memory_vs_throughput_103B}
        \small MLP 103B, batch size 256}
        \end{subfigure}
        \vspace{-0.1in}
    \caption{
        \label{fig:mlp_memory_vs_throughput}
        Peak per-device memory vs throughput for MLP training models with fixed batch sizes across different distribution strategies as measured in simulation. The optimal strategy varies for each model and batch size.
    }
\end{figure*}

\begin{figure*}[ht]
    \centering
    \includegraphics[width=0.75\linewidth]{figures/mlp_training_memory_vs_throughput_legend.pdf} \\
        \begin{subfigure}[b]{0.30\linewidth}
        \centering
        \includegraphics[width=1.0\columnwidth]{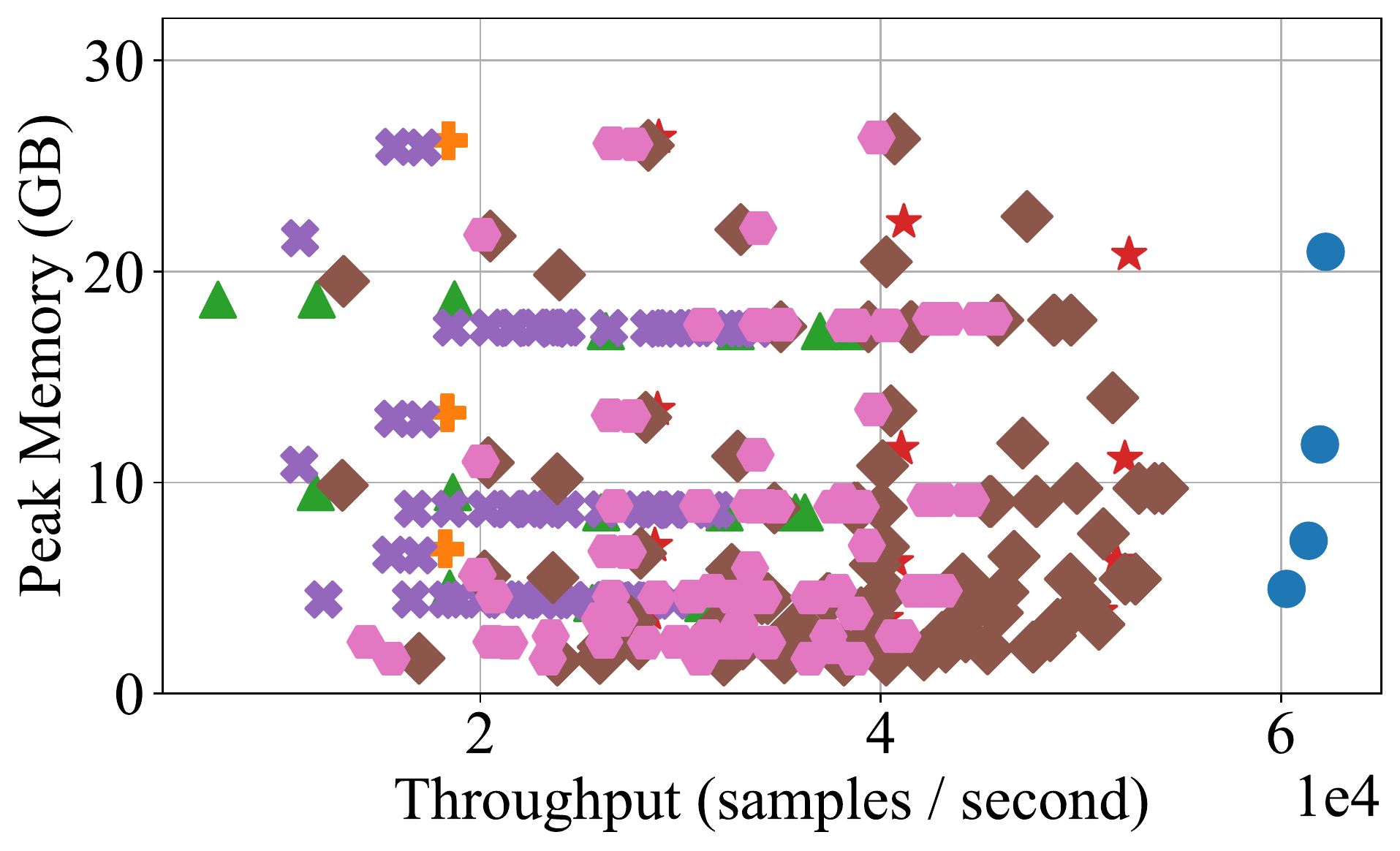}
        \caption{
        \small GPT-2 1.6B}
    \end{subfigure}
    \begin{subfigure}[b]{0.30\linewidth}
        \centering
        \includegraphics[width=1.0\columnwidth]{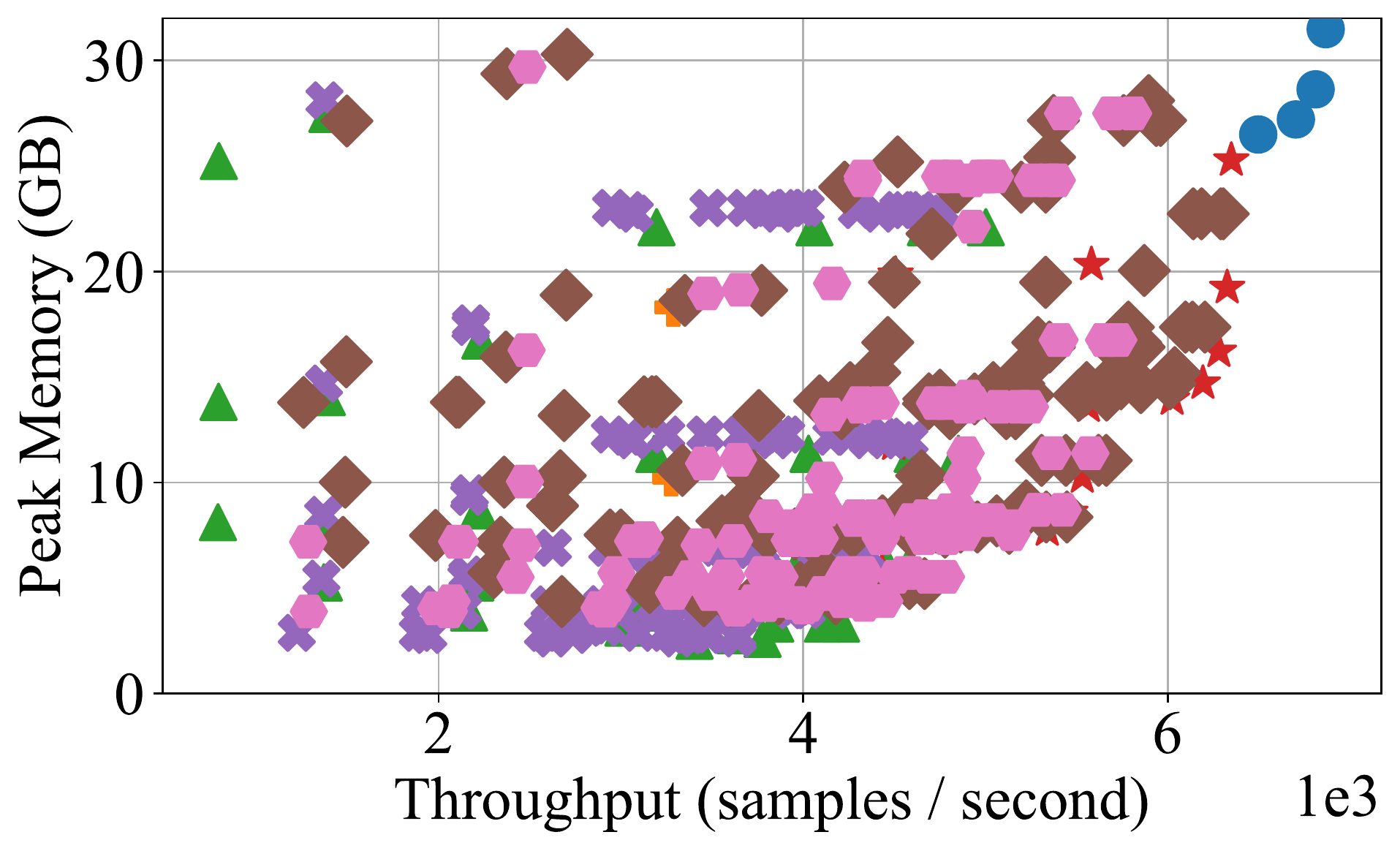}
        \caption{
        \small GPT-2 13B}
    \end{subfigure}\begin{subfigure}[b]{0.30\linewidth}
        \centering
        \includegraphics[width=1.0\columnwidth]{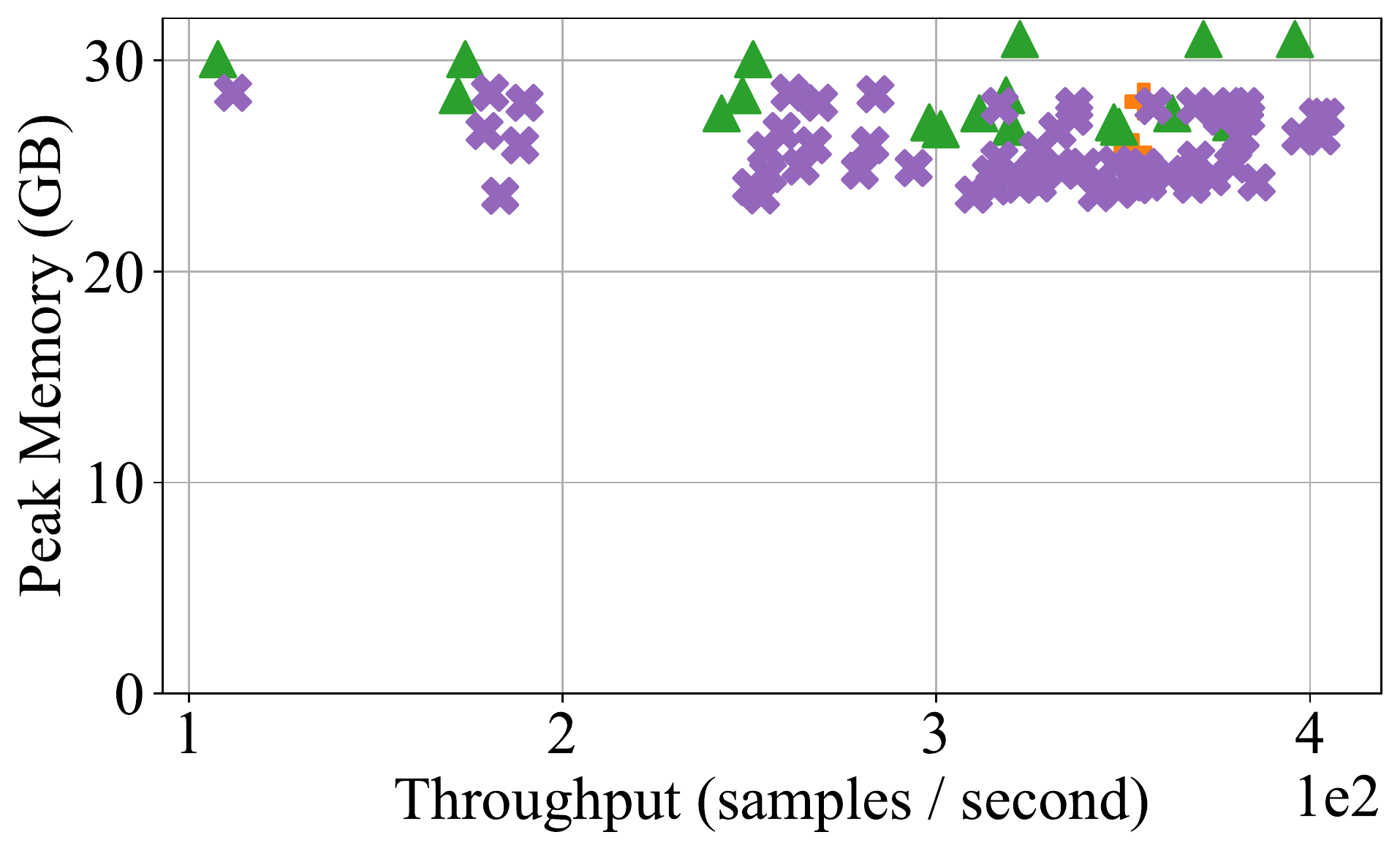}
        \caption{
        \small GPT-2 175B}
    \end{subfigure}
    \vspace{-0.1in}
    \caption{
        \label{fig:gpt2_memory_vs_throughput}
        Peak per-device memory vs throughput for GPT-2 inference models across different distribution strategies as measured in simulation. Data includes all power-of-two batch sizes between 128 and 1048576. The optimal strategy varies for each model size.
    }
    \vspace{-0.2in}
\end{figure*}

Figures~\ref{fig:mlp_memory_vs_throughput} and~\ref{fig:gpt2_memory_vs_throughput} visualize the complexity of the D/T/P distribution strategy search space for MLP training and GPT-2 inference respectively. We enumerate this search space as follows: for each model size, we use the D/T/P transforms combining data, tensor-model, and pipeline parallelism discussed in \S\ref{section:implementation} to conduct a simulated grid search over all possible power-of-two combinations of these dimensions up to 16 GPUs. We also vary the number of microbatches from 2 to 128 for pipeline parallel configurations. For training workloads we search for top configurations with a fixed global batch size, but for inference workloads we treat the global batch size as a free variable.

Note that in some cases, the optimal distribution strategy clearly matches established heuristics, but in other cases, the top strategy would not be obvious to a human analyst. For example, we see in Figure~\ref{fig:mlp_memory_vs_throughput_103B} that pure tensor-model parallelism outperforms strategies involving pipeline parallelism; this matches the recommendation from prior work that tensor-model parallelism should be maximized within a single node for models exceeding the memory capacity of a single GPU~\cite{narayanan2021efficient}. On the other hand, pure tensor-model parallelism is not viable in Figure~\ref{fig:mlp_memory_vs_throughput_17B} because the activation memory at that batch size dominates the model parameter size, and therefore the optimal configuration is a combination of all three parallelism types. This observation does not obviously map to any known heuristic, which demonstrates why automatic search is crucial.


\subsection{Automated Distribution} \label{section:automated_distribution}

In this section, we aim to validate \ir's ability to automatically identify efficient distribution strategies, and compare the optimization overhead to manual search on real hardware. We find that \ir discovers high-performance strategies across both training and inference workloads while significantly lowering the optimization cost.

\paragraph{Setup.} We take the full search space of distributed configurations from \refSec{section:search_space} and then filter out configurations that are expected to exceed the 32-GB per GPU memory limit. From the remaining list, we simulate each strategy with \ir to predict its performance, and execute the top 10 configurations in terms of predicted performance on the 16-GPU DGX node with the \ir PyTorch backend.

As a baseline, we manually execute each model and batch size configuration with each pure distribution strategy (that is, pure data, tensor-model, and pipeline parallelism).
For pipeline parallelism, we fix 128 microbatches according to a domain expert's recommended heuristic of setting $K=8P$ ($K$ = microbatches, $P$ = pipeline stages) to minimize pipeline bubbles.

\paragraph{Results.} Tables~\ref{table:mlp_optimal_configs} and~\ref{table:gpt2_optimal_configs} present the results for MLP training and GPT-2 inference respectively. For each model size, and batch size for training, we report the best performance achieved by each of the baseline distribution strategies as well as the strategies selected by the grid search. We observe that for all model size and batch size configurations, the strategies discovered by the grid search match or even exceed the performance of the baseline configurations. Moreover, the \ir simulator is able to find these strategies far quicker than manual search on physical hardware. Table~\ref{table:large_scale_configs} details the end-to-end optimization time using \ir vs exhaustively running all configurations for each model on the DGX. We see that optimization via simulation with \ir is an order of magnitude faster for certain model sizes. This confirms that \ir can effectively use simulation to drive efficient automatic optimization.

\begin{table}[ht]
\resizebox{\columnwidth}{!}{
\begin{tabular}{@{}lrrrrrr@{}}
\toprule
Model                     & \multicolumn{1}{c}{Batch size} & \multicolumn{1}{c}{\begin{tabular}[c]{@{}c@{}}Best Config\\ ($D$\ /\ $T$\ /\ $P$\ /\ $K$)\end{tabular}}  & \multicolumn{1}{c}{vs D} & \multicolumn{1}{c}{vs T} & \multicolumn{1}{c}{vs P} \\ \midrule
\multirow{12}{*}{MLP 1B}   & 128 & 1\ /\ 16\ /\ 1\ /\ 1 & 7.5$\times$ & 1.1$\times$ & 21.7$\times$ \\
& 256 & 1\ /\ 16\ /\ 1\ /\ 1 & 6.3$\times$ & 1.1$\times$ & 21.2$\times$ \\
& 512 & 1\ /\ 16\ /\ 1\ /\ 1 & 4.1$\times$ & 1.0$\times$ & 12.9$\times$ \\
& 1024 & 1\ /\ 16\ /\ 1\ /\ 1 & 2.9$\times$ & 1.0$\times$ & 9.5$\times$ \\
& 2048 & 2\ /\ 8\ /\ 1\ /\ 1 & 2.2$\times$ & 1.2$\times$ & 7.2$\times$ \\
& 4096 & 4\ /\ 4\ /\ 1\ /\ 1 & 1.7$\times$ & 1.4$\times$ & 5.1$\times$ \\
& 8192 & 4\ /\ 4\ /\ 1\ /\ 1 & 1.3$\times$ & 1.6$\times$ & 3.6$\times$ \\
& 16384 & 8\ /\ 2\ /\ 1\ /\ 1 & 1.1$\times$ & 1.9$\times$ & 2.7$\times$ \\
& 32768 & 8\ /\ 2\ /\ 1\ /\ 1 & 1.0$\times$ & 2.1$\times$ & 2.3$\times$ \\
& 65536 & 16\ /\ 1\ /\ 1\ /\ 1 & 1.0$\times$ & - & 2.3$\times$ \\
& 131072 & 16\ /\ 1\ /\ 1\ /\ 1 & 1.0$\times$ & - & 2.4$\times$ \\
& 262144 & 16\ /\ 1\ /\ 1\ /\ 1 & 1.0$\times$ & - & 2.4$\times$ \\ \midrule
\multirow{11}{*}{MLP 17B}  & 128 & 1\ /\ 16\ /\ 1\ /\ 1 & - & 1.1$\times$ & 43.8$\times$ \\
& 256 & 1\ /\ 16\ /\ 1\ /\ 1 & - & 1.0$\times$ & 35.3$\times$ \\
& 512 & 1\ /\ 16\ /\ 1\ /\ 1 & - & 1.0$\times$ & 22.4$\times$ \\
& 1024 & 1\ /\ 16\ /\ 1\ /\ 1 & - & 1.0$\times$ & 15.6$\times$ \\
& 2048 & 1\ /\ 16\ /\ 1\ /\ 1 & - & 1.0$\times$ & 10.0$\times$ \\
& 4096 & 2\ /\ 8\ /\ 1\ /\ 1 & - & 1.1$\times$ & 6.1$\times$ \\
& 8192 & 4\ /\ 4\ /\ 1\ /\ 1 & - & 1.2$\times$ & 3.7$\times$ \\
& 16384 & 4\ /\ 4\ /\ 1\ /\ 1 & - & - & 2.4$\times$ \\
& 32768 & 2\ /\ 4\ /\ 2\ /\ 8 & - & - & 1.5$\times$ \\
& 65536 & 4\ /\ 2\ /\ 2\ /\ 8 & - & - & 1.3$\times$ \\
& 131072 & 4\ /\ 2\ /\ 2\ /\ 32 & - & - & - \\ \midrule
\multirow{3}{*}{MLP 103B} & 128 & 1\ /\ 16\ /\ 1\ /\ 1 & - & 1.0$\times$ & 61.7$\times$ \\
& 256 & 1\ /\ 16\ /\ 1\ /\ 1 & - & 1.0$\times$ & - \\
& 512 & 1\ /\ 16\ /\ 1\ /\ 1 & - & 1.0$\times$ & - \\
& 1024 & 1\ /\ 16\ /\ 1\ /\ 1 & - & 1.0$\times$ & - \\ \bottomrule
\end{tabular}
}
\caption{The best distribution configuration, as predicted by the \ir grid search, for each MLP training model and batch size. We also report the speedup in throughput of this configuration against pure data, tensor-model, and pipeline parallelism baselines on physical hardware.
Note that it matches or exceeds the performance of the baselines in all cases.}
\label{table:mlp_optimal_configs}
\end{table}

\begin{table}[ht]
\resizebox{\columnwidth}{!}{
\begin{tabular}{@{}lrrrrr@{}}
\toprule
Model      & \multicolumn{1}{c}{\begin{tabular}[c]{@{}c@{}}Best Config\\ (Batch size\ /\ $D$\ /\ $T$\ /\ $P$\ /\ $K$)\end{tabular}}  & \multicolumn{1}{c}{vs D} & \multicolumn{1}{c}{vs T} & \multicolumn{1}{c}{vs P} \\ \midrule
GPT-2 1.6B & 32768\ /\ 16\ /\ 1\ /\ 1\ /\ 1 & 1.0$\times$ & 3.7$\times$ & 2.9$\times$ \\
GPT-2 13B  & 16384\ /\ 16\ /\ 1\ /\ 1\ /\ 1 & 1.0$\times$ & 2.6$\times$ & 2.0$\times$ \\
GPT-2 175B & 1024\ /\ 1\ /\ 8\ /\ 2\ /\ 2 & - & 1.3$\times$ & 5.6$\times$ \\
\bottomrule
\end{tabular}
}
\caption{The best distribution configuration, as predicted by the \ir grid search, for each GPT-2 inference model size and its throughput speedup versus pure data, tensor-model, and pipeline parallelism baselines on physical hardware.
The top pure baseline configurations were chosen by running all batch sizes that fit in memory starting from 128. The grid search configuration matches or exceeds the performance of the pure baselines in all cases.}
\label{table:gpt2_optimal_configs}
\end{table}

\subsection{Simulator Accuracy} \label{section:simulator_accuracy}

\ir's simulator aims to to accurately rank the performance of distributed configurations on physical hardware. To measure this accuracy, we first randomly sample 70+ distributed configurations from the same space as in \refSec{section:search_space} for each of the 6 model architectures and execute these configurations on the DGX using the \ir PyTorch backend. We then compare the simulated throughput with the throughput measured on real hardware. We also compute the Spearman correlation coefficient~\cite{spearman1961proof} between the simulated and real throughput values; this value captures the similarity in ranking order between two variables, so we can apply it here to determine the effectiveness of \ir's ranking methodology.

Table~\ref{table:simulator_accuracy_correlation} presents the results. We observe strong correlation for both MLP training and GPT-2 inference for all model sizes. However, there are still gaps between the absolute throughput values measured in simulation vs on physical hardware (see Figure~\ref{fig:simulator_accuracy} in the Appendix for more details). We attribute these discrepancies to the fact that we only use regression-based cost functions for few ops and use heuristics for the rest (\refSec{section:implementation}); future work will include using profiled costs to improve raw throughput prediction accuracy.

Furthermore, our memory estimation is sometimes inaccurate because our backend allocates memory naively when needed, which leads to fragmentation and out-of-memory errors for configurations that the simulator predicts will fit on the device.
Since \ir determines all ops and their schedule explicitly, standard ahead-of-time allocation strategies would avoid such fragmentation.

\begin{table}[ht]
\centering
\begin{tabular}{@{}lrrr@{}}
\toprule
\multicolumn{1}{c}{\multirow{2}{*}{Model}} & \multicolumn{1}{c}{\multirow{2}{*}{$N$}} & \multicolumn{2}{c}{Correlation}               \\ \cmidrule(l){3-4} 
\multicolumn{1}{c}{}                       & \multicolumn{1}{c}{}                   & \multicolumn{1}{c}{$r$} & \multicolumn{1}{c}{$p$} \\ \midrule
MLP 1B                                     & 94                                     & .97                   & $< 10^{-56}$                      \\
MLP 17B                                    & 94                                     & .98                   & $< 10^{-62}$                      \\
MLP 103B                                   & 71                                     & .99                   & $< 10^{-64}$                      \\ \midrule
GPT-2 1.6B                                 & 94                                     & .98                   & $< 10^{-62}$                      \\
GPT-2 13B                                  & 75                                     & .94                   & $< 10^{-34}$                      \\
GPT-2 175B                                 & 78                                     & .84                   & $< 10^{-21}$                      \\ \bottomrule
\end{tabular}
\caption{Spearman correlation coefficients as a measure of simulator ranking accuracy. Configurations were selected uniformly randomly from the simulated grid search results and then run on the 16-GPU DGX node.}
\label{table:simulator_accuracy_correlation}
\end{table}

\subsection{Simulator Scalability} \label{section:scalability}

A key property of \ir is that it enables fast simulation, because all scheduling decisions are directly embedded in the IR. In this section, we test this claim in practice.

\begin{figure}[t]
        \centering
        \includegraphics[width=1.0\columnwidth]{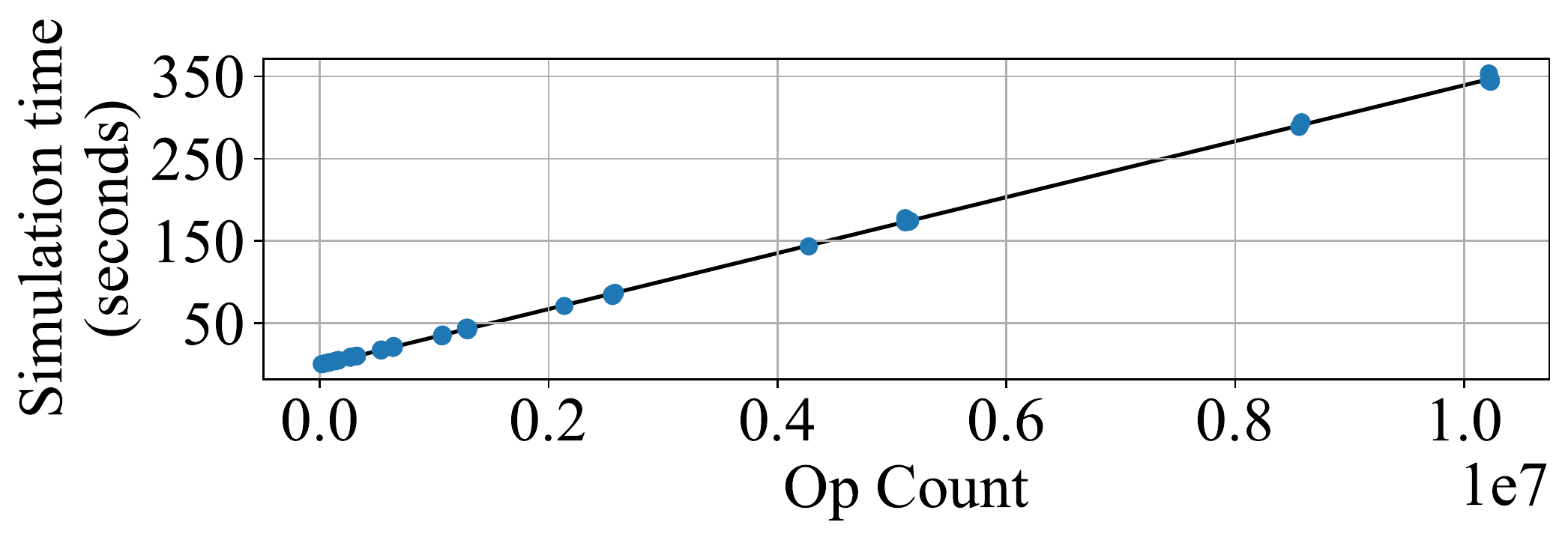}
    \caption{
        \label{fig:simulator_scaling}
        Scaling behavior of \ir simulator. We observe linear scaling with respect to the program op count.
    }
\end{figure}

Figure~\ref{fig:simulator_scaling} demonstrates how the \ir simulator scales with respect to the program op count. We measure the wall clock execution time taken to simulate a sample of distributed GPT-2 models drawn from the search space in \refSec{section:search_space} and observe linear scaling as a function of the op count.

We note that the raw simulation times would improve significantly from a compiled (e.g. C++) implementation, but this is orthogonal to our core contributions.

\section{Related Work} \label{section:related_work}


\paragraph{Eager Frameworks.}
Many existing libraries for distributed DNNs~\cite{narayanan2019pipedream, shoeybi2019megatron, rasley2020deepspeed} are implemented in eager frameworks such as PyTorch~\cite{paszke2019pytorch} that lack an IR.
They allow unpredictable dynamic behavior which makes it extremely difficult to write analyses such as a general-purpose simulator.
PipeDream~\cite{narayanan2019pipedream} uses cost models to optimize the partitioning of a model, but these are specific to their pipeline-parallel strategy. 

\paragraph{SPMD IRs.}
Graph-based frameworks such as XLA~\cite{leary2017xla} and ONNX Runtime~\cite{onnxruntime} have IRs that have been extended to represent distributed computation~\cite{yu2018dynamic, huang2019gpipe, lepikhin2021gshard, fedus2021switch, xu2021gspmd}.
These works use the single-program-multiple-data (SPMD) methodology, because it provides concise representations of common data-parallel programs.
In \ir, we can outline such repetitive blocks of code into functions to reduce IR size (so far we have not needed to, see \refSec{section:scalability}).
On the other hand, as discussed in \refSec{section:dist_ir}, SPMD-based frameworks have trouble representing pipeline parallelism within the IR.

\paragraph{Other IRs.}
PartIR~\cite{vytiniotis2020partir} is an IR for partitioning tensor programs that is useful for high-level transformations, but the distribution (mapping of computation blocks to devices) is determined by a later lowering step.
One could import the lowered program into \ir in order to integrate with our simulator.




DaCe~\cite{ben2019stateful}, Lift~\cite{steuwer2017lift}, and Elevate~\cite{hagedorn2020language} all propose IRs for representing parallel computation.
However, these IRs are primarily designed for maximizing single-node parallelism rather than optimizing distributed performance for large-scale DNNs.
Halide~\cite{ragan2017halide} separates what is computed from how it is computed; we plan to investigate integrating Halide's approach in \ir in order to make transforms more modular.
TVM~\cite{chen2018tvm} is an end-to-end optimizing compiler for DNNs, but to our knowledge it does not consider distribution.



\paragraph{DNN profiling and simulation.} 
DayDream~\cite{zhu2020daydream} and DNNMem~\cite{gao2020estimating} propose profiler-based simulators to accurately predict DNN execution time and memory respectively.
However, DayDream has a fixed set of primitives for expressing optimizations and DNNMem operates over front-end model specifications, while we capture both model and distribution in a generic IR.
FlexFlow~\cite{jia2019beyond} uses a simulator to search over a fixed strategy space but does not consider pipeline parallelism. Similarly, PipeDream-2BW~\cite{narayanan2020memory} includes a profiler to predict performance for various pipeline-parallel configurations but does not consider horizontal parallelism. 
\ir's simulator is more general as it is not tied to a particular class of models or strategies.

\section{Future Work} \label{section:discussion}

There are three promising directions for future work.

First, one can upgrade our grid search to more sophisticated algorithms for automatic distribution, such as MCMC search~\cite{jia2019beyond}, integer and dynamic programming~\cite{narayanan2019pipedream, tarnawski2020efficient}, reinforcement learning~\cite{wang2020auto, mirhoseini2017device}, and custom algorithms~\cite{narayanan2020memory, jia2019taso}.
Most of these are complementary to \ir, as we can use \ir's simulator as their cost functions, and we plan to investigate these.

Second, we plan to integrate \ir with popular distribution frameworks in order to support more DNN models and distribution strategies.
The quick option is to use \ir as shown in \refSec{section:evaluation} to predict the best distributed configuration $(D, T, P, K)$, and feed that to, e.g., DeepSpeed~\cite{rasley2020deepspeed}, for execution.
However, one would have to do extra work to make sure that the transforms implemented in \ir stay in sync with the transforms implemented in DeepSpeed for the predictions to remain optimal.
Alternatively, IR-based frameworks such as Jax/XLA can adopt \ir as their representation of distribution.
This would require porting the distribution strategies to be \ir transforms, but this can potentially simplify their implementation as the lowering pass can be reused.
In either case, one must empirically tune the cost models (as shown in \refSec{section:implementation}) so that the simulator matches the backend.

Finally, we can improve our simulator's runtime and memory estimations.
Runtime estimation can be improved by extending empirical op cost models (\refSec{section:implementation}) for all ops, and by using test inputs that correspond to input shapes seen during execution of real models.
We can improve memory estimation by accounting for the temporary memory used by each op during its execution, which can be estimated empirically or analytically for ops with known kernels.



\section{Conclusion} \label{section:conclusion}

\ir is an efficient IR for explicit representation of distributed DNN computation.
\ir permits efficient static analyses such as simulation that accelerate manual distribution as well as enable automatic distribution via search algorithms.
Expressing distribution as transformations over \ir functions allows one to develop hybrid strategies via composition of existing strategies.
We demonstrate how \ir can be used to facilitate automatic distribution by finding optimal strategies for large models among a hybrid space of distributions.

\bibliography{paper}
\bibliographystyle{mlsys2022}

\clearpage

\appendix

\section{Appendix}

\subsection{Expressivity Examples}

\begin{figure}[]
\begin{lstlisting}[%
  emph={[1]D1, x, y, w1, w2, as_f, p, dp, as_b, dw2, das, dw1, w1_new, w2_new},
  emphstyle={[1]\color{\getDeviceColor{1}}},
%   emph={[2]D2, y, w2, y_1, ar_1, y_2, ar_2, das_1, dw2_1, das_2, dw2_2, dw2, w2_new, p_1, p_2, dp_1, dp_2},
%   emphstyle={[2]\color{\getDeviceColor{2}}},
  firstnumber=1,
  % moredelim=**[is][\color{\getDeviceColor{1}}]{|}{|},
  % moredelim=**[is][\color{\getDeviceColor{2}}]{|(}{)|},
  ]

func @mlp_GradientCheckpointing(
    %w1: D1,
    %w2: D1,
    %x: D1,
    %y: D1) {
  %as_f: D1 = @dense(%w1, %x)
  %p: D1 = @dense(%w2, %as_f) |\label{line-p}|
  // Memory for %as_1 is reclaimed
  %dp: D1 = @lossGrad(%p, %y)
  // Recompute activation
  %as_b: D1 = @dense(%w1, %x) |\label{line-as_b}|
  %dw2, %das: D1 = @denseGrad(%w2, %as_b, %dp)
  %dw1, _: D1 = @denseGrad(%w1, %x, %das)
  // Weight update (WU)
  %w1_new: D1 = Optimizer(%w1, %dw1)
  %w2_new: D1 = Optimizer(%w2, %dw2)
  return %w1_new, %w2_new
 }
}

\end{lstlisting}
\caption{\ir code listing for sequential training of a 2-layer MLP model using gradient checkpointing~\cite{chen2016training}.
}
  \label{fig:gradient_checkpointing}
\end{figure}

\begin{figure}[]
\begin{lstlisting}[%
  emph={[1]D1, w1, x_1, y_1, as_1, w2_1_f, p_1, dp_1, w2_1_b, dw2_1, das_1, dw1_1, dw1, w1_new},
  emphstyle={[1]\color{\getDeviceColor{1}}},
  emph={[2]D2, w2, x_2, y_2, w1_2_f, as_2, p_2, dp_2, dw2_2, das_2, w1_2_b, dw1_2, dw2, w2_new},
  emphstyle={[2]\color{\getDeviceColor{2}}},
  firstnumber=1,
  % moredelim=**[is][\color{\getDeviceColor{1}}]{|}{|},
  % moredelim=**[is][\color{\getDeviceColor{2}}]{|(}{)|},
  ]

func @mlpDP_ZeRO(
    %w1: D1,
    %w2: D2,
    %x_1: D1,
    %x_2: D2,
    %y_1: D1,
    %y_2: D2) {
  %as_1: D1 = @dense(%w1, %x_1) |\label{line_as_1}|
  // %w1 is managed exclusively by device 1
  %w1_2_f: D2 = Send(%w1, 2)    |\label{line_w1_2_1}|
  %as_2: D2 = @dense(%w1_2_f, %x_2)
  // Memory for %w1_2_f is reclaimed
  // %w2 is managed exclusively by device 2
  %w2_1_f: D1 = Send(%w2, 1)    |\label{line_w2_1_1}|
  %p_1: D1 = @dense(%w2_1_f, %as_1) |\label{line_p_1}|
  // Memory for %w2_1_f is reclaimed
  %p_2: D2 = @dense(%w2, %as_2)
  %dp_1: D1 = @lossGrad(%p_1, %y_1)
  %dp_2: D2 = @lossGrad(%p_2, %y_2)
  // %w2_1_f has been freed, so re-send %w2
  %w2_1_b: D1 = Send(%w2, 1)
  %dw2_1, %das_1: D1 = @denseGrad(%w2_1_b, %as_1, %dp_1)
  %dw2_2, %das_2: D2 = @denseGrad(%w2, %as_2, %dp_2)
  %dw1_1, _: D1 = @denseGrad(%w1, %x_1, %das_1)
  // %w1_2_f has been freed, so re-send %w1
  %w1_2_b: D2 = Send(%w1, 2)
  %dw1_2, _: D2 = @denseGrad(%w1_2_b, %x_2, %das_2)
  // Weight update (WU)
  %dw1: D1 = MPIReduce([%dw1_1, %dw1_2], 1) |\label{line_dw1}|
  %dw2: D2 = MPIReduce([%dw2_1, %dw2_2], 2) |\label{line_dw2}|
  %w1_new: D1 = Optimizer(%w1, %dw1)
  %w2_new: D2 = Optimizer(%w2, %dw2)
  return %w1_new, %w2_new
 }
}

\end{lstlisting}
\caption{\ir code listing for data-parallel training of a 2-layer MLP model over 2 devices with ZeRO-style parameter and gradient partitioning (ZeRO stages 2 and 3)~\cite{rajbhandari2019zero}.
}
  \label{fig:zero}
\end{figure}
\begin{figure}[]
    \centering
    \includegraphics[width=0.85\columnwidth]{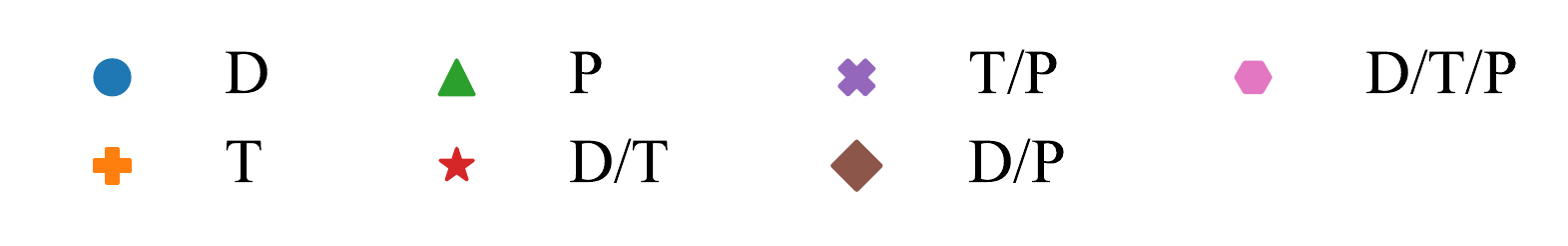} \\
    \begin{subfigure}[b]{0.5\columnwidth}
        \centering
        \includegraphics[width=1.0\columnwidth]{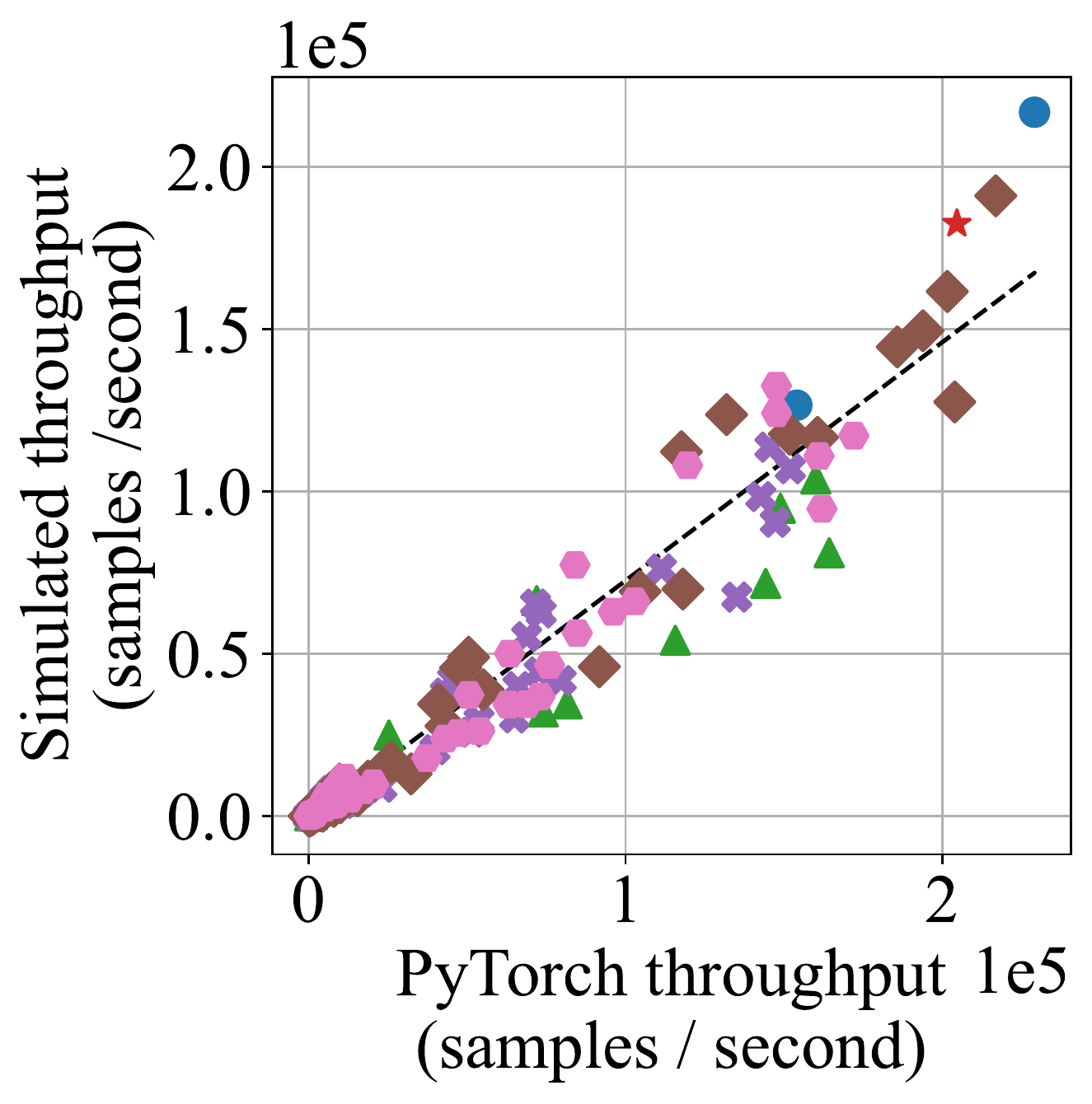}
        \caption{
        \small MLP training.}
        \label{fig:mlp_simulator_accuracy}
    \end{subfigure}
    \begin{subfigure}[b]{0.48\columnwidth}
        \centering
        \includegraphics[width=1.0\columnwidth]{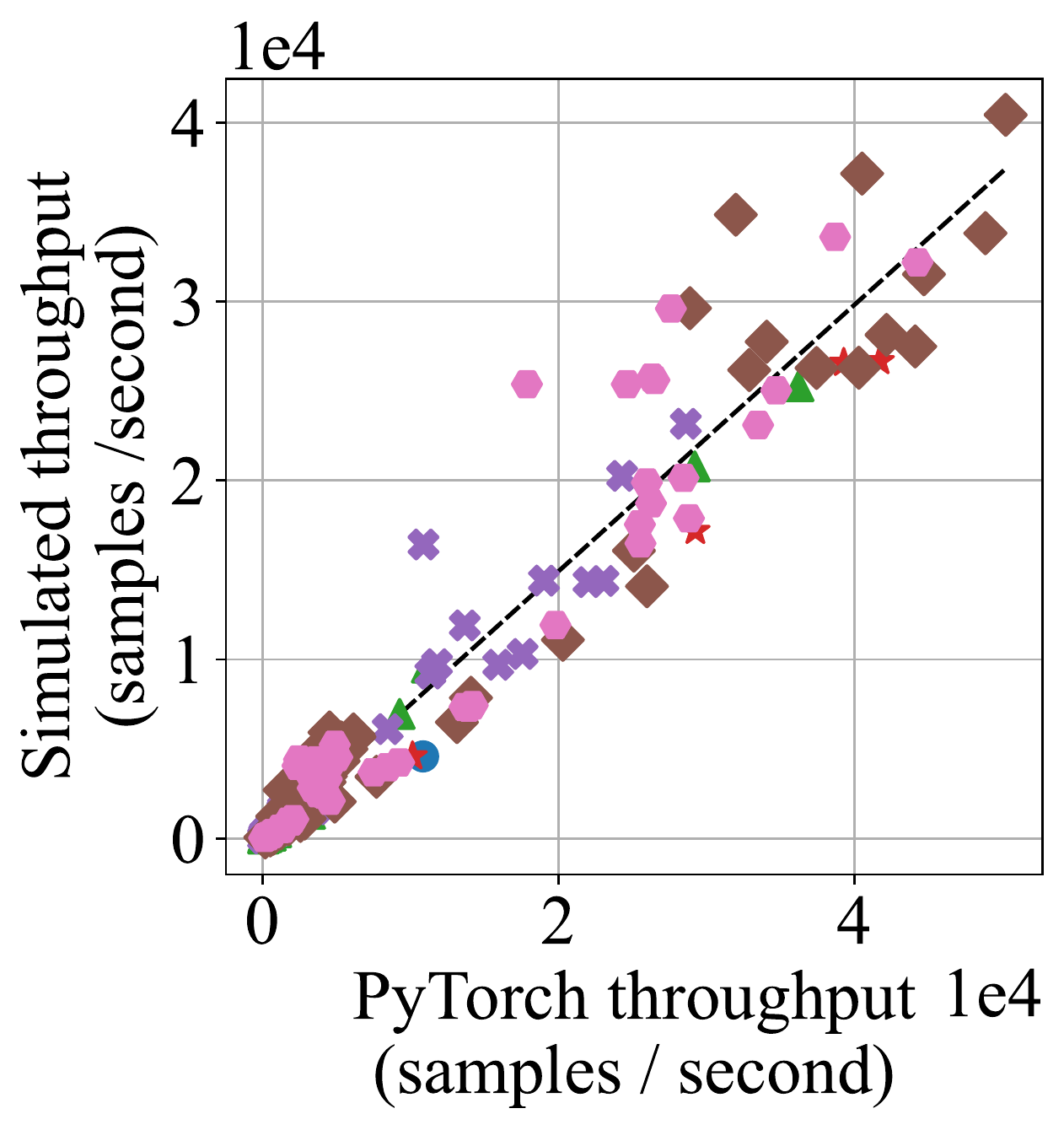}
        \caption{
        \small GPT-2 inference.}
        \label{fig:gpt2_simulator_accuracy}
    \end{subfigure}
    \vspace{-0.1in}
    \caption{
        Simulated performance predicted by the \ir simulator vs real performance measured by the \ir backend. 
    }
    \label{fig:simulator_accuracy}
    \vspace{-0.2in}
\end{figure}

In this section we provide additional examples to highlight \ir's expressivity. In particular, \refFig{fig:gradient_checkpointing} demonstrates gradient checkpointing~\cite{chen2016training} and \refFig{fig:zero} demonstrates ZeRO partitioning~\cite{rajbhandari2019zero}.

Gradient checkpointing is a memory-saving optimization which entails temporarily discarding activations in the forward pass after certain checkpoint nodes have finished executing (thereby reclaiming their memory) and then re-computing these activations in the backward pass when they are needed to compute the relevant gradients. This improves upon the default memory usage pattern which keeps all activations in device memory throughout the entire duration of the forward pass. \refFig{fig:gradient_checkpointing} presents an example of gradient checkpointing using \ir. In this program, the first activation ($as_f$) is discarded after line~\ref{line-p} and is then re-computed at line~\ref{line-as_b} in the backward pass ($as_b$).

The ZeRO partitioning algorithms eliminate redundant memory usage in data-parallel training by distributing optimizer state (stage 1), gradients (stage 2), and parameters (stage 3) across nodes. Figure~\ref{fig:zero} provides an example of ZeRO stages 2 and 3 in a \ir program\footnote{\ir can also represent ZeRO stage 1 given a fine-grained specification of optimizer state, but we limit optimizer details in our examples for brevity.}. In this example, $w_{1}$ and its gradient are assigned exclusively to device 1, while $w_{2}$ and its gradient are assigned exclusively to device 2. Therefore $w_{1}$ must be sent to device 2 in line~\ref{line_w1_2_1} so that device 2 can execute the first dense layer. Similarly $w_{2}$ must be sent to device 1 in line~\ref{line_w2_1_1} for executing the second dense layer. The \texttt{MPIReduce} calls on lines~\ref{line_dw1} and~\ref{line_dw2} aggregate the gradients for $w_{1}$ and $w_{2}$ on devices 1 and 2 respectively. 

\subsection{Simulator Accuracy}

Figure~\ref{fig:simulator_accuracy} compares the real throughput achieved by distributed MLP and GPT-2 configurations using the \ir backend against the throughput predicted by the \ir simulator for the same configurations. We generally find that the simulator produces accurate predictions of raw throughput, but there are cases where the error is more pronounced. As discussed in Section~\ref{section:simulator_accuracy}, we attribute these cases to the current lack of profiled op costs. Furthermore, we observe that distributed configurations involving pipeline parallelism tend to incur more prediction error; we suspect this is a result of the non-uniform communication patterns inherent to pipeline parallel execution. More fine-grained profiling could mitigate such discrepancies. 



\end{document}